\pdfoutput=1

\documentclass[11pt]{article}

\usepackage[preprint]{acl}
\usepackage{amsmath}

\usepackage{times}
\usepackage{latexsym}
\usepackage{amsfonts}
\usepackage{algorithm}
\usepackage{algorithmic}

\newcommand\C[1]\null
\usepackage[T1]{fontenc}
\usepackage{dblfloatfix}

\usepackage{tipa}

\usepackage{listings}
\usepackage{xcolor}

\usepackage[utf8]{inputenc}

\usepackage{microtype}

\usepackage{inconsolata}

\usepackage{graphicx}

\usepackage{float}
\title{Cross-cultural Adaptation of Recipes for Low-Resource Languages: \\ A Benchmark and Multilingual Culinary Lexicon for \\ Hausa, Yorùbá, and Igbo}

\author{
\begin{tabular}{c}
Nuhu Ibrahim, Maryam Abdulazeez, Fatsuma Jauro, Salisu Aliyu,\\ Robert Stevens, Riza Batista-Navarro
\end{tabular}
\\[0.5em]
Department of Computer Science, The University of Manchester, United Kingdom
\\
\textbf{Correspondence:} 
\texttt{nuhu.ibrahim@manchester.ac.uk}
}

\usepackage{multirow}

\begin{document}
\maketitle
\begin{abstract}
Recent multilingual large language models (LLMs) have improved machine translation, yet culturally adaptive translation for low-resource languages (LRLs) remains underexplored. This is particularly challenging for recipes, where ingredients, cooking tools, measurements, and procedures may require adaptation to the target cultural context. In this work, we introduce ReciLoReL, a benchmark of 600 expert-translated and culturally adapted recipes across Yor\`ub\'a, Hausa, and Igbo (three of the most prominent languages in Nigeria), and ReciLex, a multilingual culinary lexicon containing approximately 15,000 translated entities across ten recipe entity categories, including ingredients, tools, chef actions, state of tools, and durations. We evaluate state-of-the-art multilingual LLMs for English-to-LRL recipe translation under zero-shot and few-shot settings and investigate whether augmenting LLMs with ReciLex improves culturally adaptive recipe translation without requiring large target-language recipe corpora for training or fine-tuning. Beyond conventional reference-based translation evaluation using BLEU, chrF, and TER, we introduce five criteria for evaluating cultural adaptation, adapted from established cross-cultural adaptation methods. Across the selected LLMs and languages, ReciLex improves average BLEU from 13.57 to 14.51 and chrF from 37.89 to 38.74, while reducing TER from 77.14 to 75.12. Human evaluation further shows improvements ranging from 0.11 to 0.36 points across the five cultural adaptation evaluation criteria on a five-point scale. Overall, our work provides new resources, evaluation criteria, and empirical evaluation for studying culturally adaptive recipe translation in low-resource settings.
\end{abstract}

\section{Introduction}
Recent advances in multilingual LLMs and machine translation (MT) systems have improved translation quality for many high-resource languages \cite{garcia2023unreasonable, hendy2023good, robinson2023chatgpt}. However, performance remains considerably weaker for low-resource languages (LRL), particularly in culturally grounded domains where translation requires more than direct lexical mapping \cite{zhu2024multilingual, iyer2024quality}. While recent work has begun exploring culturally adaptive recipe translation and generation \cite{cao2024cultural, hu2024bridging, hu2025culinary}, existing approaches primarily focus on high-resource settings, such as Chinese--English adaptation, and often rely on existing large target-language recipe corpora.

To the best of our knowledge, culturally adaptive recipe translation has not been explored for low-resource African languages such as Yor\`ub\'a, Hausa, and Igbo. Recipe translation often requires cultural adaptation in addition to semantic preservation. Ingredients, utensils, measurements, and cooking procedures commonly used in one culture may be unavailable, unfamiliar, or linguistically ambiguous in another, requiring models to generate translations that are both culturally meaningful and practically feasible. Evaluating such adaptation also presents a challenge, as conventional machine translation metrics primarily measure lexical or surface-form similarity to reference translations and do not directly assess whether the resulting recipe fits the target cultural context.

To address this gap, we introduce a benchmark for culturally adaptive recipe translation across three low-resource Nigerian languages: Yor\`ub\'a, Hausa, and Igbo. We construct a gold-standard dataset of 600 manually translated and culturally adapted recipes derived from the RecipeNLG corpus~\cite{bien2020recipenlg}. The benchmark supports evaluation of English-to-Yor\`ub\'a, English-to-Hausa, and English-to-Igbo translations, measuring not only reference-based translation performance but also the cultural adaptation of the generated recipes. To support this evaluation, we draw on established cross-cultural adaptation approaches in healthcare and existing work on recipe cultural adaptation to define five evaluation criteria: ingredient localisation, procedural localisation, semantic preservation, cultural authenticity, and linguistic naturalness (see Section~\ref{sec:cultural-evaluation}). We evaluate these criteria using human evaluation.

We further investigate whether lightweight multilingual lexical supervision can improve culturally grounded translation without requiring access to large-scale parallel corpora. Using the ReciFine corpus~\cite{ibrahim2026recifine}, we construct a multilingual culinary lexicon by translating the top 15{,}000 most frequent entities across ten recipe entity categories. During inference, we retrieve relevant translated entities and provide them to the LLM as auxiliary guidance for recipe cultural adaptation. Our findings suggest that translating a relatively small number of carefully selected lexical entities can substantially improve low-resource recipe translation without the cost of constructing large translated parallel corpora. All code and datasets associated with this work are released publicly under CC BY-NC 4.0\footnote{\href{https://github.com/nuhu-ibrahim/ReciLoReL-ReciLex}{https://github.com/nuhu-ibrahim/ReciLoReL-ReciLex}}.

Overall, our contributions are as follows:
\begin{itemize}
    \item We introduce a benchmark for evaluating culturally adaptive recipe translation in Yor\`ub\'a, Hausa, and Igbo based on 600 expert-translated and culturally adapted recipes.

    \item We introduce five criteria for evaluating cultural adaptation in recipe translation, adapted from established cross-cultural adaptation approaches, and apply them through human evaluation.

    \item We present a multilingual culinary lexicon containing 15{,}000 expert-translated entities to support culturally adaptive recipe translation for LRLs.

    \item We demonstrate that lexicon augmentation improves culturally adaptive recipe translation in LRLs across both reference-based translation and cultural adaptation evaluation.
\end{itemize}

\section{Related Work}
\label{sec:related-work}
\subsection{Cross-Cultural Adaptation}
Cross-cultural adaptation considers both linguistic translation and the cultural changes required when transferring content from one setting to another. This distinction has been extensively studied in the healthcare domain, particularly for adapting health-related quality of life (HRQoL) instruments and questionnaires~\citep{guillemin1993cross,herdman1998model} and self-report measures~\citep{beaton2000guidelines}. \citet{guillemin1993cross} proposed guidelines for the cross-cultural adaptation of HRQoL measures, considering semantic, idiomatic, experiential, and conceptual equivalence between source and target versions. \citet{herdman1998model} further proposed a model for examining different forms of equivalence when adapting HRQoL instruments across cultures. Building on this work, \citet{beaton2000guidelines} described cross-cultural adaptation as a process that considers both language and cultural adaptation when preparing an instrument for use in another setting.

Importantly, translation and cultural adaptation are not always required together. \citet{guillemin1993cross} show that a language change requires translation, while a change in cultural setting may require cultural adaptation even where the language remains unchanged. When both language and cultural setting change, both translation and cultural adaptation are required. This distinction matters for recipe cultural adaptation to LRLs, where recipes cross both linguistic and culinary-cultural contexts. Beyond healthcare, recent NLP work has similarly explored cultural adaptation of general text~\cite{peskov2021adapting,singh2024translating,vijaya2025cross}, further highlighting the need to account for cultural context when transferring text between communities.

\subsection{Cross-Cultural Recipe Adaptation}
Cross-cultural adaptation is particularly relevant to recipes because ingredients, measurements, cooking tools, techniques, and recipe-writing conventions can vary considerably across cultures. \citet{cao2024cultural} introduced cross-cultural recipe adaptation between Chinese and English and showed that recipe adaptation requires more than direct translation. They introduced CulturalRecipes, a Chinese--English dataset constructed by pairing recipes for similar dishes across the two cultures, and evaluated machine translation models and LLMs. Their analysis identifies differences in ingredients, measurement units, cooking tools, techniques, and recipe styles, showing the different forms cultural adaptation can take in recipe text.

Subsequent work has approached cross-cultural recipe adaptation through retrieval. CARROT~\cite{hu2024bridging} formulates the problem as cross-cultural recipe retrieval, where recipes from a target-culture collection are retrieved while balancing preservation of the source recipe with appropriateness to the target culture. The framework uses culturally aware query rewriting and re-ranking to improve retrieval across Chinese and English recipes. CARRIAGE~\cite{hu2025culinary} extends retrieval-based adaptation through retrieval-augmented generation, using recipes from the target cuisine as context for generating culturally appropriate adaptations.

These approaches provide important evidence that target-culture recipes can support cultural adaptation, but they address a different resource setting from ours. CARROT requires a collection of recipes from the target culture because its objective is to retrieve an appropriate target-culture recipe, while CARRIAGE similarly uses target-culture recipes as retrieval context for generation. Such resources are difficult to obtain for many low-resource languages, including Yor\`ub\'a, Hausa, and Igbo. Rather than retrieving complete target-culture recipes, our work investigates whether a considerably smaller multilingual culinary lexicon can provide useful cultural and linguistic guidance during translation. The approaches are therefore complementary rather than directly comparable: retrieval-based methods assume the availability of target-culture recipe collections, whereas our setting investigates adaptation where such collections are unavailable.

This setting is also related to work in low-resource and multilingual machine translation. Multilingual models continue to show weaker performance for many African languages \cite{adelani2023masakhanews,bhat2023large,ibrahim2025large}, particularly where domain-specific and culturally grounded terminology is required. Previous work has shown that translation dictionaries, terminology-aware supervision, and retrieval augmentation can improve multilingual translation and generation~\cite{lu2024chain,yong2024lexc,chang2025enhancing}. Motivated by these findings, we investigate whether translated culinary entities can provide a lightweight alternative to the large target-language resources used by existing recipe cultural adaptation approaches.

\subsection{Evaluating Cultural Adaptation}
Evaluating cross-cultural recipe adaptation presents a different problem from evaluating direct translation. \citet{cao2024cultural} use reference-based metrics including BLEU, chrF, ROUGE-L, and BERTScore, but note that multiple valid adaptations can exist for the same source recipe. They therefore complement automatic evaluation with human evaluation based on four criteria: grammar, consistency, source-recipe preservation, and cultural appropriateness. Their results also show that cultural appropriateness correlates weakly with automatic metrics, highlighting the difficulty of measuring cultural adaptation with reference-based metrics alone.

CARROT~\cite{hu2024bridging} similarly considers both preservation of the original recipe and cultural appropriateness to the target culture when evaluating cross-cultural recipe retrieval. This reflects an important property of the task: a culturally adapted recipe should be appropriate for the target culture while retaining the essential characteristics of the source recipe.

While these studies provide approaches for evaluating recipe cultural adaptation, cultural appropriateness remains relatively broad. In this work, we provide a more fine-grained evaluation of the specific properties affected by cultural adaptation. We draw on the recipe cultural adaptation literature together with established cross-cultural adaptation approaches from healthcare \cite{guillemin1993cross,herdman1998model,beaton2000guidelines} to define five criteria for recipe cultural adaptation: ingredient localisation, procedural localisation, semantic preservation, cultural authenticity, and linguistic naturalness. These criteria are described in Section~\ref{sec:cultural-evaluation}.

\section{Cross-Cultural Recipe Adaptation}
\subsection{Task Formulation}
\label{sec:task}
We define \textit{cross-cultural recipe adaptation} as the translation of a recipe from a source language and cultural context to a target language and cultural context, while adapting culturally dependent recipe elements where necessary. Given a source recipe $R_s$ in language $L_s$ and culinary context $C_s$, the task is to generate a target recipe $R_t$ in language $L_t$ and culinary context $C_t$: 

\begin{math} f(R_s, L_s, C_s) \rightarrow (R_t, L_t, C_t). \end{math}

The task combines two related objectives: translating the recipe from $L_s$ to $L_t$ and, where necessary, adapting it from $C_s$ to $C_t$. However, cultural adaptation does not imply unrestricted modification of the source recipe. Recipe elements that transfer appropriately to the target context are preserved, while adaptation is applied where direct transfer would make an element unfamiliar, inaccessible, unnatural, or impractical. We define the scope of recipe cultural adaptation in Table~\ref{tab:adaptation-scope}.

\begin{table}[ht!]
\centering
\scriptsize
\resizebox{\columnwidth}{!}{
\begin{tabular}{|l|l|l|}
\hline
\textbf{Element} & \textbf{Adapt when} & \textbf{Preserve} \\
\hline
Ingredients &
Unavailable/unfamiliar &
Culinary function \\
\hline
Measurements &
Convention differs &
Quantity/proportion \\
\hline
Tools/equipment &
Unavailable/uncommon &
Intended function \\
\hline
Methods/procedures &
Impractical/unfamiliar &
Preparation logic \\
\hline
Culinary expressions &
Unnatural when translated &
Intended meaning \\
\hline
\end{tabular}
}
\caption{Scope and constraints of cultural adaptation in recipe translation, adapted from cross-cultural adaptation principles in \citet{guillemin1993cross} and \citet{cao2024cultural}.}
\label{tab:adaptation-scope}
\end{table}

\subsection{Evaluating Cultural Adaptation}
\label{sec:cultural-evaluation}
Cultural adaptation presents a different evaluation problem from conventional machine translation evaluation. Metrics such as BLEU, chrF, and TER primarily measure lexical or surface-form similarity between a generated translation and a gold reference. However, appropriate cultural adaptation may replace an ingredient, measurement, utensil, cooking procedure, or culinary expression (e.g., ``cook until fork-tender'') with forms that differ from the surface form of a single gold reference. Thus, we evaluate cultural adaptation separately from reference-based translation performance.

To define the evaluation criteria, we draw on established work on cross-cultural adaptation, primarily from the healthcare and recipe domains~\citep{guillemin1993cross,herdman1998model,beaton2000guidelines,cao2024cultural}. Appendix~\ref{appendix:cultural_evaluation_guideline} (Table~\ref{tab:cultural-dimensions}). The five criteria are independently assessed using a five-point Likert scale to measure different aspects of cultural adaptation in recipe translation.

\paragraph{Ingredient Localisation (INGR):}
This evaluates whether ingredients are familiar, culturally appropriate, and reasonably available in the target context. It is informed by \textit{experiential equivalence}, which considers whether an item or experience represented in the source is applicable in the target cultural setting~\cite{guillemin1993cross,beaton2000guidelines}. For recipes, this includes determining whether source ingredients can be retained directly or whether culturally appropriate alternatives are required.

\paragraph{Procedural Localisation (PROC):}
This evaluates the appropriateness of cooking methods, equipment, measurements, and preparation conventions in the target context. It similarly draws on \textit{experiential equivalence}~\cite{guillemin1993cross,beaton2000guidelines}, as a cooking procedure may be correctly translated but remain difficult to interpret or perform using practices and resources common to the target community.

\paragraph{Semantic Preservation (SEM):}
This measures whether the adapted recipe retains the source recipe's culinary meaning and intended outcome. It follows \textit{semantic equivalence} and \textit{preservation}, which concern preserving meaning across source and target versions~\cite{guillemin1993cross,beaton2000guidelines,cao2024cultural}. This criterion also constrains the extent of cultural adaptation, i.e., substitutions or procedural changes should not alter the recipe to the point that its original culinary intent is lost.

\paragraph{Cultural Authenticity (CULT):}
This evaluates whether the resulting recipe is culturally plausible within the target context. It is informed by \textit{conceptual equivalence} and \textit{cultural appropriateness}, which consider whether the underlying concept represented in the source remains meaningful in the target culture~\cite{guillemin1993cross,beaton2000guidelines,cao2024cultural}. For recipe cultural adaptation, this concerns whether the adapted ingredients, procedures, and their combination form a recipe appropriate to the target culinary setting.

\paragraph{Linguistic Naturalness (LING):}
This evaluates whether the culinary terminology and instructional expressions of the resulting recipe read naturally in the target language. It is informed by \textit{idiomatic equivalence}, which addresses expressions that cannot be appropriately transferred through literal translation~\cite{guillemin1993cross,beaton2000guidelines}. A culturally adapted recipe should therefore use expressions and recipe-writing conventions that are natural to speakers of the target language.

\section{Benchmark and Lexicon Construction}
In this work, we introduce two complementary resources for culturally adaptive recipe translation in low-resource languages (LRLs): ReciLoReL, a multilingual benchmark dataset for culturally adaptive recipe translation, and ReciLex, a multilingual culinary lexicon spanning ten recipe entity categories, including ingredients, tools, chef actions, state of tools and durations. Together, these resources support evaluating and improving culturally grounded recipe adaptation across Yorùbá, Hausa, and Igbo. ReciLoReL contains 600 expert-translated and culturally adapted recipes translated from English into each of the three aforementioned languages. Beyond literal translation, annotators adapted recipes to reflect culturally relevant ingredients, measurements, cooking tools, and preparation methods commonly used within the target communities. This enables evaluation of both translation quality and cultural adaptation in English-to-LRL settings. In addition, we introduce ReciLex, a multilingual culinary lexicon constructed from the most frequent entities from the ReciFine corpus \cite{ibrahim2026recifine}. The entities were originally identified through named entity recognition (NER) over more than 2.2 million recipes using methods described in \citet{ibrahim2026knowledge}, ensuring that the translated entities reflect commonly occurring culinary concepts.

\subsection{ReciLoReL: Our Benchmark Dataset}
\label{section:benchmark_dataset}
To construct ReciLoReL, we selected 200 recipes from the RecipeNLG corpus \cite{bien2020recipenlg} using a controlled sampling procedure (see Appendix~\ref{appendix:recilorel_recipe_selection}) designed to balance recipe complexity and ensure coverage of culinary entities across all ten ReciFine entity type categories (see Appendix~\ref{appendix:recifine_et_cats}). We then randomly sampled recipes that met these criteria and translated them from English into Yor\`ub\'a, Hausa, and Igbo, resulting in 600 culturally adapted recipe translations.

The benchmark supports English-to-LRL recipe cultural adaptation evaluation for all three languages. Annotators had formal knowledge of their respective languages and a basic understanding of cooking and food within the local context. They were instructed to preserve the meaning and culinary intent of each recipe while adapting ingredients, cooking utensils, measurements, and preparation procedures as needed for the target community. 

Annotation guidelines were iteratively refined and are provided in Appendix~\ref{appendix:recilorel_recipe_translation_guideline}. To ensure annotation quality, annotators initially translated five sample recipes, which were independently reviewed by three experts per language. We discussed common issues and used them to refine the guidelines. Expert review was also used to identify suspected MT- or LLM- generated annotations. When an expert suspected MT or LLM use, other experts independently reviewed the annotator's translations. We removed the annotator only after reviewers agreed and the annotator confirmed that MT or LLM had been used. In total, only two annotators were removed and their translations re-annotated. This process ensured that translations were not excluded solely because of a particular translation style.

Overall, the benchmark was constructed using two Hausa annotators, five Yor\`ub\'a annotators, and two Igbo annotators, alongside three expert reviewers per language for quality validation.  Table~\ref {tab:recilorel_dataset_statistics} in Appendix~\ref {appendix:desc_stats_recilorel} presents descriptive statistics for ReciLoReL, including cross-language lexical and structural characteristics, translation expansion ratios, and annotation details.

\subsection{ReciLex: Our Culinary Lexicon}
We further introduce ReciLex, a multilingual culinary lexicon designed to support lexicon-augmented recipe cultural adaptation in low-resource settings. From ReciFine, we selected the top 500 most frequent entities from each of the ten culinary entity categories, including ingredients, chef actions, tools, measurements, durations, etc. (see Appendix~\ref{appendix:recifine_et_cats}). Expert annotators then manually translated these entities into Yorùbá, Hausa, and Igbo, resulting in a multilingual culinary lexicon of approximately 15{,}000 translated entities. Using high-frequency entities ensures that ReciLex captures commonly occurring culinary concepts. To ensure annotation quality, each annotator initially translated 20 entities from each of the 10 entity categories, which were independently reviewed by three language experts per language. Common annotation issues were discussed collectively to refine the annotation guidelines (see Appendix~\ref{appendix:recilex_lexicon_translation_guideline}), while annotators suspected of consistently producing low-quality or automatically generated translations were excluded from the final annotation process using similar approaches discussed in Section~\ref{section:benchmark_dataset}. Table~\ref {tab:recilex_statistics} in Appendix~\ref {appendix:desc_stats_recilex} presents descriptive statistics for ReciLex, including multilingual lexicon coverage, lexical characteristics, translation expansion ratios, and annotation details.

\section{Experimental Setup}
\subsection{Baseline (Without ReciLex): Adaptation by LLM Prompting and MT Models}
\label{baseline:adaptation_by_llms_mt_models}
We evaluate multiple multilingual LLMs under zero-shot and few-shot prompting settings for culturally adaptive recipe translation. In the zero-shot setting, models were prompted to translate and culturally adapt recipes without additional examples. For few-shot prompting, demonstrations were sampled from the ReciLoReL validation split and provided as in-context examples during generation. We selected the number of demonstrations based on preliminary experiments reported in Appendix~\ref{appendix:fewshot_comparison}. We additionally evaluate dedicated multilingual MT models as translation baselines. 

\subsection{ReciLex: Adaptation using our Multilingual Lexicon}
To investigate whether lightweight multilingual lexicon augmentation improves culturally adaptive recipe translation, we augment zero-shot and few-shot prompting experiments using translated culinary lexicons from ReciLex. For each recipe, entities are first identified by matching against the ReciLex multilingual culinary lexicons. Matching translated entities are then retrieved and used to augment the prompt as contextual guidance during generation. 

\subsection{Multilingual LLMs and MT Models}
We evaluate a diverse set of multilingual LLMs spanning proprietary, open-source, and African-focused multilingual systems. Specifically, we evaluate Gemma-27B, Llama-3.3-70B, and GPT-5.1 due to their strong multilingual capabilities and recent use in multilingual and low-resource NLP research~\cite{ustun2024aya,ojo2025afrobench}. We additionally evaluate NLLB-200~\citep{costa2022no} and M2M-100~\citep{fan2021beyond} as dedicated multilingual MT baselines to provide a reference for direct translation performance. Appendix~\ref {app:model_configurations} provides details on the evaluated models, including their parameter sizes, inference backends, quantisation settings, and experimental configurations. 

\subsection{Evaluation Metrics}
We evaluate translation quality using BLEU \cite{papineni2002bleu}, TER \cite{snover2006study} and chrF \cite{popovic2015chrf}. We additionally evaluate cultural adaptation using the five criteria defined in Section~\ref{sec:cultural-evaluation} through human evaluation. Appendix~\ref {appendix:cultural_evaluation_guideline} provides the annotation guidelines used to assess cultural adaptation.

\section{Results and Discussion}
\subsection{Automatic Translation Evaluation}
Table~\ref{tab:main_results} presents English-to-Yor\`ub\'a, Hausa, and Igbo recipe translation performance with and without ReciLex augmentation. Results are aggregated across GPT-5.1, Llama-70B, and Gemma-27B, combining zero-shot and few-shot English prompting to provide an overall comparison of translation performance for each language. ReciLex improves automatic evaluation scores across BLEU, chrF, and TER for Hausa and Igbo, but yields lower scores for Yor\`ub\'a. This divergence suggests that lexicon augmentation's effectiveness varies across languages and highlights the potential limitations of reference-based metrics in evaluating culturally adaptive translation. Notably, the decline in automatic scores for Yor\`ub\'a contrasts with improvements observed in human evaluations of cultural adaptation (Table~\ref{tab:cultural_results}), suggesting that automatic metrics may not fully capture culturally appropriate lexical choices and linguistic variation. Appendix~\ref{sec:orthographic_analysis} investigates the possible reasons for the lower Yor\`ub\'a scores following ReciLex augmentation.

\begin{table}[ht!]
\centering
\scriptsize
\setlength{\tabcolsep}{4pt}
\begin{tabular}{|l|ccc|ccc|}
\hline
\multirow{2}{*}{\textbf{Language}} &
\multicolumn{3}{c|}{\textbf{Without ReciLex}} &
\multicolumn{3}{c|}{\textbf{ReciLex}} \\
\cline{2-7}
& \textbf{BLEU}$\uparrow$
& \textbf{chrF}$\uparrow$
& \textbf{TER}$\downarrow$
& \textbf{BLEU}$\uparrow$
& \textbf{chrF}$\uparrow$
& \textbf{TER}$\downarrow$ \\
\hline
Yor\`ub\'a
& \textbf{8.14} & \textbf{29.48} & \textbf{83.93}
& 2.36 & 17.23 & 96.63 \\
Hausa
& 16.29 & 43.55 & 71.03
& \textbf{21.28} & \textbf{52.76} & \textbf{58.81} \\
Igbo
& 16.28 & 40.64 & 76.43
& \textbf{19.90} & \textbf{46.24} & \textbf{69.93} \\
\hline
\end{tabular}
\caption{Aggregated English-to-Yor\`ub\'a, Hausa, and Igbo recipe translation performance. Scores are averaged equally across GPT-5.1, Llama-70B, and Gemma-27B under zero-shot and few-shot prompting, with and without ReciLex. Bold indicates the better score.}
\label{tab:main_results}
\end{table}

\subsection{Manual Cultural Adaptation Evaluation}
\label{sec:cultural_adaptation}
Since automatic translation metrics do not directly capture cultural appropriateness, we conducted a human evaluation across five dimensions (see Section~\ref{sec:cultural-evaluation}).

We randomly selected 80 recipes and evaluated their translations into Yor\`ub\'a, Hausa, and Igbo using GPT-5.1, Llama-3.3-70B, and Gemma-27B, yielding 2,880 translation outputs that were manually evaluated altogether (Appendix~\ref{tab:annotation_distribution}). Each recipe was translated once per model under zero-shot and few-shot English prompting, with and without ReciLex. Ratings were assigned on a five-point scale and averaged across models and languages.

Table~\ref{tab:cultural_results} shows that ReciLex improves human evaluation scores across all dimensions for Yor\`ub\'a and Igbo, whereas Hausa exhibits mixed results, with improvements in ingredient localisation and cultural authenticity.

\begin{table}[ht!]
\scriptsize
\setlength{\tabcolsep}{2pt}
\begin{tabular}{|l|l|c|c|c|c|c|}
\hline
\textbf{Language} &
\textbf{Method} &
\textbf{INGR} &
\textbf{PROC} &
\textbf{SEM} &
\textbf{CULT} &
\textbf{LING} \\
\hline

\multirow{2}{*}{Yor\`ub\'a}
& Without ReciLex
& 3.29 & 3.17 & 3.23 & 2.48 & 2.98 \\
& ReciLex
& \textbf{4.13} & \textbf{3.55} & \textbf{3.69}
& \textbf{3.03} & \textbf{3.34} \\
\hline

\multirow{2}{*}{Hausa}
& Without ReciLex
& 2.94 & \textbf{2.88} & \textbf{3.07}
& 2.97 & \textbf{2.68} \\
& ReciLex
& \textbf{3.07} & 2.72 & 2.93
& \textbf{3.05} & 2.66 \\
\hline

\multirow{2}{*}{Igbo}
& Without ReciLex
& 2.81 & 3.03 & 2.61 & 2.33 & 2.58 \\
& ReciLex
& \textbf{2.93} & \textbf{3.15} & \textbf{2.73}
& \textbf{2.41} & \textbf{2.67} \\
\hline

\end{tabular}
\caption{Mean human evaluation scores with and without ReciLex for Yor\`ub\'a, Hausa, and Igbo, averaged across three LLMs and combining zero-shot and few-shot English prompting. Scores are on a five-point scale (higher is better). Bold indicates the better score.}
\label{tab:cultural_results}
\end{table}

Qualitative analysis further demonstrates ReciLex's benefits in improving culinary terminology and reducing translation errors. For instance, ReciLex corrects the mistranslation of \emph{fish} as \emph{wild animal} in Igbo, replaces untranslated expressions such as \emph{sour cream} in Hausa, and produces more recognisable Yor\`ub\'a terminology for \emph{shrimp}. It also reduces English code-mixing, improving lexical localisation and linguistic naturalness. Nevertheless, some procedural inaccuracies remain, particularly in cooking methods and durations. Expanding ReciLex's coverage of culinary actions, techniques, and temporal expressions may provide richer lexical guidance and help address these limitations. Appendix~\ref{sec:qualitative-error-analysis} provides further illustrative examples, including English source recipes, their translations with and without ReciLex in the target languages, and detailed analyses of the observed improvements and errors.

Inter-annotator agreement (IAA) was assessed using graded soft agreement, yielding average scores of 0.630, 0.735, and 0.850 for Yor\`ub\'a, Hausa, and Igbo, respectively.  Detailed IAA results and the soft agreement methodology are provided in Appendix~\ref{app:cultural_agreement}, while a breakdown of annotation coverage and the number of recipes evaluated for each reported human cultural adaptation result is presented in Appendix~\ref{sec:annotation_coverage}.

\subsection{Ablation Study}
We conduct ablation studies to examine the effects of ReciLex augmentation, prompting strategy, terminology realisation, and orthographic variation on translation quality and cultural adaptation. Appendix~\ref {sec:ablation_studies} provides additional analyses of language- and model-specific performance, dedicated MT baselines, prompt language, and few-shot example selection.


\subsubsection{Overall Impact of ReciLex}
\label{sec:ablation_recilex}
To assess ReciLex's overall impact, we aggregate zero-shot and few-shot results across GPT-5.1, Llama-70B, and Gemma-27B for English-to-Yor\`ub\'a, Hausa, and Igbo recipe translation.

Tables~\ref{tab:recilex_auto_impact} and \ref{tab:recilex_human_impact} show that ReciLex improves overall automatic translation performance, increasing BLEU and chrF while reducing TER. Human evaluation similarly reveals improvements across all five cultural adaptation dimensions, particularly ingredient localisation and cultural authenticity. These findings demonstrate ReciLex's effectiveness in supporting culturally grounded recipe translation.

\begin{table}[ht!]
\centering
\scriptsize
\setlength{\tabcolsep}{5pt}
\begin{tabular}{|l|r|r|r|}
\hline
\textbf{Setting} &
\textbf{BLEU $\uparrow$} &
\textbf{chrF $\uparrow$} &
\textbf{TER $\downarrow$} \\
\hline
Without ReciLex & 13.57 & 37.89 & 77.14 \\
ReciLex & \textbf{14.51} & \textbf{38.74}
& \textbf{75.12} \\
\hline
\end{tabular}
\caption{Mean automatic evaluation scores across three LLMs, three target languages, and zero-shot and few-shot prompting. Bold indicates the better score.}
\label{tab:recilex_auto_impact}
\end{table}

\begin{table}[ht!]
\centering
\small
\setlength{\tabcolsep}{3pt}
\begin{tabular}{|l|r|r|r|r|r|}
\hline
\textbf{Setting} &
\textbf{INGR} &
\textbf{PROC} &
\textbf{SEM} &
\textbf{CULT} &
\textbf{LING} \\
\hline
Without ReciLex
& 3.01 & 3.03 & 2.97 & 2.59 & 2.75 \\
ReciLex
& \textbf{3.37} & \textbf{3.14}
& \textbf{3.12} & \textbf{2.83}
& \textbf{2.89} \\
\hline
\end{tabular}
\caption{Mean human cultural adaptation scores across three LLMs, three target languages, and zero-shot and few-shot prompting. Bold indicates the better score.}
\label{tab:recilex_human_impact}
\end{table}

\subsubsection{Zero-Shot vs Few-Shot Prompting with ReciLex}
\label{sec:ablation_recilex_zero_and_few_shot_impact}
As shown in Tables~\ref{tab:recilex_auto_ablation} and~\ref{tab:recilex_human_ablation}, few-shot prompting consistently outperforms zero-shot prompting under ReciLex augmentation in the aggregate, improving BLEU, chrF and TER. Human evaluation similarly shows improvements across all five cultural adaptation dimensions, particularly ingredient localisation and cultural authenticity. These findings suggest that in-context demonstrations provide additional benefits even when domain-specific lexical knowledge is available.

\begin{table}[ht!]
\centering
\scriptsize
\setlength{\tabcolsep}{5pt}
\begin{tabular}{|l|r|r|r|}
\hline
\textbf{Setting} &
\textbf{BLEU $\uparrow$} &
\textbf{chrF $\uparrow$} &
\textbf{TER $\downarrow$} \\
\hline
Zero-shot + ReciLex
& 10.70 & 34.96 & 78.78 \\
Few-shot + ReciLex
& \textbf{18.32} & \textbf{42.53} &
\textbf{71.47} \\
\hline
\end{tabular}
\caption{Automatic translation performance under zero-shot and few-shot ReciLex-augmented prompting, averaged across three LLMs and target languages. Bold indicates the better score.}
\label{tab:recilex_auto_ablation}
\end{table}

\begin{table}[ht!]
\centering
\small
\setlength{\tabcolsep}{3pt}
\begin{tabular}{|l|r|r|r|r|r|}
\hline
\textbf{Setting} &
\textbf{INGR} &
\textbf{PROC} &
\textbf{SEM} &
\textbf{CULT} &
\textbf{LING} \\
\hline
Zero-shot + ReciLex
& 3.30 & 3.11 & 3.07 & 2.78 & 2.85 \\
Few-shot + ReciLex
& \textbf{3.45} & \textbf{3.17}
& \textbf{3.16} & \textbf{2.88}
& \textbf{2.92} \\
\hline
\end{tabular}
\caption{Human cultural adaptation ratings under zero-shot and few-shot ReciLex-augmented prompting, averaged across three LLMs and target languages. Bold indicates the better score.}
\label{tab:recilex_human_ablation}
\end{table}

\subsubsection{Effect of ReciLex on Culinary Entity Realisation}
\label{subsec:recilex_entity_realisation}
To assess ReciLex's contribution to culinary entity translation, we measured how often LLM-generated translations included the target-language expressions recommended by ReciLex. Using the culinary entity categories described in Appendix~\ref{appendix:recifine_et_cats}, we calculated terminology agreement as the percentage of source entities with available ReciLex mappings for which at least one recommended translation appeared in the generated recipe. We counted each distinct entity once per recipe and category, regardless of the number of alternative translations.

As shown in Table~\ref{tab:recilex_adherence}, incorporating ReciLex substantially increased terminology agreement across all models and target languages. Under zero-shot prompting, improvements ranged from 58.87 to 78.84 percentage points, while few-shot prompting yielded gains of 44.62 to 63.18 percentage points. Overall, these findings demonstrate that the LLMs actively incorporate lexical information from ReciLex into their generated translations, confirming that the provided lexicon influences their choice of culinary expressions.

\begin{table}[ht!]
\centering
\scriptsize
\setlength{\tabcolsep}{5pt}
\renewcommand{\arraystretch}{1.1}
\begin{tabular}{|l|l|c|c|c|}
\hline
\textbf{Lang.} &
\textbf{Model} &
\textbf{Shot} &
\textbf{Without ReciLex} &
\textbf{ReciLex} \\
\hline

\multirow{6}{*}{Yor\`ub\'a}
& \multirow{2}{*}{GPT-5.1}
  & Zero-Shot & 8.84 & \textbf{79.42} \\
& & Few-Shot & 12.17 & \textbf{75.24} \\
\cline{2-5}
& \multirow{2}{*}{Llama-70B}
  & Zero-Shot & 1.90 & \textbf{80.74} \\
& & Few-Shot & 7.97 & \textbf{71.15} \\
\cline{2-5}
& \multirow{2}{*}{Gemma-27B}
  & Zero-Shot & 3.43 & \textbf{76.62} \\
& & Few-Shot & 11.82 & \textbf{72.73} \\
\hline

\multirow{6}{*}{Hausa}
& \multirow{2}{*}{GPT-5.1}
  & Zero-Shot & 24.95 & \textbf{87.17} \\
& & Few-Shot & 30.34 & \textbf{85.01} \\
\cline{2-5}
& \multirow{2}{*}{Llama-70B}
  & Zero-Shot & 14.60 & \textbf{90.49} \\
& & Few-Shot & 28.18 & \textbf{81.48} \\
\cline{2-5}
& \multirow{2}{*}{Gemma-27B}
  & Zero-Shot & 17.22 & \textbf{92.33} \\
& & Few-Shot & 30.27 & \textbf{87.96} \\
\hline

\multirow{6}{*}{Igbo}
& \multirow{2}{*}{GPT-5.1}
  & Zero-Shot & 13.62 & \textbf{72.49} \\
& & Few-Shot & 21.87 & \textbf{73.01} \\
\cline{2-5}
& \multirow{2}{*}{Llama-70B}
  & Zero-Shot & 5.75 & \textbf{73.84} \\
& & Few-Shot & 17.57 & \textbf{62.19} \\
\cline{2-5}
& \multirow{2}{*}{Gemma-27B}
  & Zero-Shot & 12.17 & \textbf{73.56} \\
& & Few-Shot & 18.67 & \textbf{67.84} \\
\hline
\end{tabular}
\caption{ReciLex terminology agreement (\%) across zero-shot and few-shot prompting. Bold indicates the better score. Results are computed over matched source entities from successfully generated recipes.}
\label{tab:recilex_adherence}
\end{table}

\subsubsection{Orthographic and Reference Sensitivity}
\label{sec:orthographic_analysis}
Despite improved terminology realisation, ReciLex yields lower reference-based scores for Yor\`ub\'a. To investigate this discrepancy, we analyse orthographic variation in baseline and ReciLex translations across all three languages, using 80 recipes per language.

Table~\ref{tab:orthographic_analysis} shows that Unicode Normalisation Form C (NFC) affects all Yor\`ub\'a ReciLex outputs, compared with only 17 zero-shot and 21 few-shot baseline outputs, while Hausa and Igbo exhibit no or negligible changes. Combining marks also increase substantially in Yor\`ub\'a ReciLex translations. Given the importance of diacritics in Yor\`ub\'a, these differences may affect reference-based evaluation. Additionally, valid lexical alternatives, including synonyms, borrowings, and paraphrases, may differ from reference translations despite being linguistically appropriate.

Further analysis in Appendix~\ref{app:yoruba_normalisation} shows that orthographic normalisation substantially improves automatic evaluation scores for Yor\`ub\'a.

\begin{table}[ht!]
\centering
\scriptsize
\setlength{\tabcolsep}{1.5pt}
\begin{tabular}{|l|l|c|c|c|c|}
\hline
\textbf{Language} &
\textbf{Prompt} &
\multicolumn{2}{c|}{\textbf{NFC changes}} &
\multicolumn{2}{c|}{\textbf{Marks/100}} \\
\cline{3-6}
& & \textbf{Without ReciLex} & \textbf{ReciLex}
& \textbf{Without ReciLex} & \textbf{ReciLex} \\
\hline
Hausa & Zero-Shot & 0/80 & 0/80 & 0.00 & 0.00 \\
Hausa & Few-Shot & 0/80 & 0/80 & 0.00 & 0.00 \\
\hline
Igbo & Zero-Shot & 3/80 & 2/80 & 0.03 & 0.00 \\
Igbo & Few-Shot & 0/80 & 0/80 & 0.00 & 0.00 \\
\hline
Yor\`ub\'a & Zero-Shot & 17/80 & 80/80 & 1.96 & 11.30 \\
Yor\`ub\'a & Few-Shot & 21/80 & 80/80 & 2.44 & 9.39 \\
\hline
\end{tabular}
\caption{Cross-language orthographic characteristics of GPT-5.1 translations. NFC changes denote recipes modified by Unicode canonical normalisation; Marks/100 denotes combining marks per 100 Unicode code points.}
\label{tab:orthographic_analysis}
\end{table}

\section{Conclusion}
We introduced ReciLoReL, a benchmark for culturally adaptive English-to-Yor\`ub\'a, Hausa, and Igbo recipe translation, and ReciLex, a multilingual culinary lexicon spanning ten recipe entity categories. Experiments across multilingual LLMs demonstrate that ReciLex augmentation generally improves automatic translation metrics for Hausa and Igbo, while yielding more consistent gains in human evaluations of cultural adaptation, particularly ingredient localisation and cultural authenticity across the languages. The divergence between automatic and human evaluation results for Yor\`ub\'a further highlights the limitations of reference-based metrics in assessing cultural appropriateness, consistent with \citet{cao2024cultural}. Unlike approaches requiring large target-language recipe corpora~\citep{hu2024bridging,hu2025culinary}, ReciLex provides a lightweight, lexicon-based mechanism for improving culturally grounded translation in low-resource settings.

Further analyses show that few-shot prompting generally outperforms zero-shot prompting, while English-language prompts yield modest improvements over target-language prompts. ReciLex also increases the realisation of recommended culinary terminology in generated recipes, demonstrating its influence on lexical choices. Moreover, LLMs generally outperform dedicated MT models in both translation quality and cultural adaptation.

\section{Limitations}
Although ReciLoReL provides a manually curated benchmark for culturally adaptive recipe translation, its relatively limited size reflects the substantial effort required to produce and validate culturally appropriate translations. While prioritising annotation quality, this limits the diversity of recipes and culinary practices represented. Future work could expand the benchmark to support more comprehensive evaluation across a wider range of culinary contexts.

Our experiments focus on Yor\`ub\'a, Hausa, and Igbo, enabling targeted investigation of lexicon-augmented translation for three under-resourced African languages collectively spoken by over 200 million people. However, this linguistic scope limits the generalisability of our findings. Extending ReciLoReL and ReciLex to additional low-resource languages would help establish whether the observed benefits of lexical augmentation transfer across different linguistic structures and cultural contexts.

Finally, the divergence between automatic and human evaluation results, particularly for Yor\`ub\'a, highlights broader challenges in evaluating low-resource language generation. Future research should investigate more robust evaluation methods that capture semantic equivalence and valid linguistic variation beyond surface-level reference similarity, particularly where language-specific evaluation resources remain limited.

\bibliography{custom}
\appendix

\section{Supplementary Ablation Studies}
\label{sec:ablation_studies}

\subsection{Language and Model Performance}
\label{sec:ablation_language}
We examine automatic translation quality and human judgements of cultural adaptation separately for each target language and model.

\subsubsection{Yor\`ub\'a}
\label{sec:ablation_yoruba}
Tables~\ref{tab:yoruba_auto_ablation} and \ref{tab:yoruba_human_ablation} reveal contrasting effects of ReciLex augmentation. Across all three LLMs, ReciLex reduces automatic translation scores, with lower BLEU and chrF and higher TER. However, human evaluation shows consistent improvements across all five cultural adaptation dimensions and all models, particularly ingredient localisation and cultural authenticity. These findings suggest that reference-based metrics may not fully capture culturally appropriate adaptations introduced through lexical augmentation.

\begin{table}[ht!]
\centering
\scriptsize
\setlength{\tabcolsep}{4pt}
\begin{tabular}{|l|l|r|r|r|}
\hline
\textbf{Model} &
\textbf{Setting} &
\textbf{BLEU $\uparrow$} &
\textbf{chrF $\uparrow$} &
\textbf{TER $\downarrow$} \\
\hline
\multirow{2}{*}{GPT-5.1}
& Without ReciLex & \textbf{11.65} & \textbf{34.65} & 83.70 \\
& ReciLex & 4.05 & 19.70 & 98.40 \\
\hline
\multirow{2}{*}{Llama-70B}
& Without ReciLex & 5.15 & 24.85 & 86.05 \\
& ReciLex & 1.95 & 17.50 & 94.35 \\
\hline
\multirow{2}{*}{Gemma-27B}
& Without ReciLex & 7.60 & 28.95 & \textbf{82.05} \\
& ReciLex & 1.05 & 14.50 & 97.15 \\
\hline
\end{tabular}
\caption{Automatic evaluation of English-to-Yor\`ub\'a recipe translation with and without ReciLex, averaged across zero-shot and few-shot prompting. Bold indicates the better score.}
\label{tab:yoruba_auto_ablation}
\end{table}

\begin{table}[ht!]
\centering
\scriptsize
\setlength{\tabcolsep}{3pt}
\begin{tabular}{|l|l|r|r|r|r|r|}
\hline
\textbf{Model} &
\textbf{Setting} &
\textbf{INGR} &
\textbf{PROC} &
\textbf{SEM} &
\textbf{CULT} &
\textbf{LING} \\
\hline

\multirow{2}{*}{GPT-5.1}
& Without ReciLex
& 3.68 & 3.60 & 3.87 & 3.04 & 3.44 \\
& ReciLex
& \textbf{4.28} & \textbf{3.85} &
\textbf{4.10} & \textbf{3.34} &
\textbf{3.63} \\
\hline

\multirow{2}{*}{Llama-70B}
& Without ReciLex
& 3.04 & 2.88 & 2.85 & 2.12 & 2.72 \\
& ReciLex
& 4.10 & 3.39 & 3.50 & 2.94 & 3.20 \\
\hline

\multirow{2}{*}{Gemma-27B}
& Without ReciLex
& 3.16 & 3.03 & 2.98 & 2.27 & 2.78 \\
& ReciLex
& 4.00 & 3.42 & 3.47 & 2.81 & 3.18 \\
\hline

\end{tabular}
\caption{Human cultural adaptation ratings for English-to-Yor\`ub\'a translation, reported separately for each LLM and averaged across zero-shot and few-shot prompting. Bold indicates the better score.}
\label{tab:yoruba_human_ablation}
\end{table}

\subsubsection{Hausa}
\label{sec:ablation_hausa}
Tables~\ref{tab:hausa_auto_ablation} and \ref{tab:hausa_human_ablation} show that ReciLex improves automatic translation performance across all three LLMs, increasing BLEU and chrF while reducing TER. Human evaluation, however, reveals model-dependent effects. ReciLex improves all five cultural adaptation dimensions for Llama-70B, while Gemma-27B benefits in ingredient localisation, semantic preservation, and cultural authenticity. In contrast, GPT-5.1 achieves its highest human evaluation scores without ReciLex, suggesting that its existing Hausa language capabilities may reduce its reliance on external lexical guidance. These findings indicate that ReciLex is particularly beneficial for models with weaker baseline performance, helping them produce more culturally appropriate translations.

\begin{table}[ht!]
\centering
\scriptsize
\setlength{\tabcolsep}{4pt}
\begin{tabular}{|l|l|r|r|r|}
\hline
\textbf{Model} &
\textbf{Setting} &
\textbf{BLEU $\uparrow$} &
\textbf{chrF $\uparrow$} &
\textbf{TER $\downarrow$} \\
\hline
\multirow{2}{*}{GPT-5.1}
& Without ReciLex & 13.30 & 43.75 & 76.80 \\
& ReciLex & 19.45 & 52.30 & 60.95 \\
\hline
\multirow{2}{*}{Llama-70B}
& Without ReciLex & 16.75 & 41.40 & 69.40 \\
& ReciLex & \textbf{22.90} & \textbf{53.25} & \textbf{58.15} \\
\hline
\multirow{2}{*}{Gemma-27B}
& Without ReciLex & 18.80 & 45.50 & 66.90 \\
& ReciLex & 21.50 & 52.75 & 57.30 \\
\hline
\end{tabular}
\caption{Automatic evaluation of English-to-Hausa recipe translation with and without ReciLex, averaged across zero-shot and few-shot prompting. Bold indicates the better score.}
\label{tab:hausa_auto_ablation}
\end{table}

\begin{table}[ht!]
\centering
\scriptsize
\setlength{\tabcolsep}{3pt}
\begin{tabular}{|l|l|r|r|r|r|r|}
\hline
\textbf{Model} &
\textbf{Setting} &
\textbf{INGR} &
\textbf{PROC} &
\textbf{SEM} &
\textbf{CULT} &
\textbf{LING} \\
\hline

\multirow{2}{*}{GPT-5.1}
& Without ReciLex
& \textbf{3.43} & \textbf{3.51} &
\textbf{3.68} & \textbf{3.33} &
\textbf{3.21} \\
& ReciLex
& 3.03 & 2.72 & 2.92 & 3.07 & 2.72 \\
\hline

\multirow{2}{*}{Llama-70B}
& Without ReciLex
& 2.50 & 2.23 & 2.56 & 2.60 & 2.06 \\
& ReciLex
& 3.04 & 2.69 & 2.90 & 3.01 & 2.58 \\
\hline

\multirow{2}{*}{Gemma-27B}
& Without ReciLex
& 2.88 & 2.91 & 2.98 & 2.98 & 2.77 \\
& ReciLex
& 3.13 & 2.74 & 2.99 & 3.06 & 2.67 \\
\hline

\end{tabular}
\caption{Human cultural adaptation ratings for English-to-Hausa translation, reported separately for each LLM and averaged across zero-shot and few-shot prompting. Bold indicates the better score.}
\label{tab:hausa_human_ablation}
\end{table}

\subsubsection{Igbo}
\label{sec:ablation_igbo}
Tables~\ref{tab:igbo_auto_ablation} and \ref{tab:igbo_human_ablation} show that ReciLex improves automatic translation performance across all three LLMs, increasing BLEU and chrF while reducing TER. Human evaluation also shows improvements across all five cultural adaptation dimensions, particularly ingredient localisation, procedural localisation, and semantic preservation. These findings demonstrate ReciLex's benefits for English-to-Igbo recipe translation.

\begin{table}[ht!]
\centering
\scriptsize
\setlength{\tabcolsep}{4pt}
\begin{tabular}{|l|l|r|r|r|}
\hline
\textbf{Model} &
\textbf{Setting} &
\textbf{BLEU $\uparrow$} &
\textbf{chrF $\uparrow$} &
\textbf{TER $\downarrow$} \\
\hline
\multirow{2}{*}{GPT-5.1}
& Without ReciLex & 16.45 & 43.10 & 81.25 \\
& ReciLex & 20.25 & 47.10 & 71.90 \\
\hline
\multirow{2}{*}{Llama-70B}
& Without ReciLex & 14.40 & 37.55 & 75.90 \\
& ReciLex & 18.90 & 45.00 & 69.70 \\
\hline
\multirow{2}{*}{Gemma-27B}
& Without ReciLex & 18.00 & 41.30 & 72.20 \\
& ReciLex & \textbf{20.55} & \textbf{46.60} &
\textbf{68.20} \\
\hline
\end{tabular}
\caption{Automatic evaluation of English-to-Igbo recipe translation with and without ReciLex, averaged across zero-shot and few-shot prompting. Bold indicates the better score.}
\label{tab:igbo_auto_ablation}
\end{table}

\begin{table}[ht!]
\centering
\scriptsize
\setlength{\tabcolsep}{3pt}
\begin{tabular}{|l|l|r|r|r|r|r|}
\hline
\textbf{Model} &
\textbf{Setting} &
\textbf{INGR} &
\textbf{PROC} &
\textbf{SEM} &
\textbf{CULT} &
\textbf{LING} \\
\hline

\multirow{2}{*}{GPT-5.1}
& Without ReciLex
& 3.06 & 3.20 & 2.79 & 2.44 & 2.73 \\
& ReciLex
& \textbf{3.16} & \textbf{3.37} &
\textbf{2.91} & \textbf{2.60} &
\textbf{2.88} \\
\hline

\multirow{2}{*}{Llama-70B}
& Without ReciLex
& 2.54 & 2.82 & 2.40 & 2.19 & 2.39 \\
& ReciLex
& 2.70 & 3.04 & 2.58 & 2.29 & 2.58 \\
\hline

\multirow{2}{*}{Gemma-27B}
& Without ReciLex
& 2.83 & 3.09 & 2.64 & 2.35 & 2.62 \\
& ReciLex
& 2.94 & 3.05 & 2.70 & 2.35 & 2.56 \\
\hline

\end{tabular}
\caption{Human cultural adaptation ratings for English-to-Igbo translation, reported separately for each LLM and averaged across zero-shot and few-shot prompting. Bold indicates the better score.}
\label{tab:igbo_human_ablation}
\end{table}

\subsection{Yor\`ub\'a Orthographic Normalisation}
\label{app:yoruba_normalisation}
Following the orthographic differences identified in Section~\ref{sec:orthographic_analysis}, we re-evaluate ReciLex-augmented Yor\`ub\'a translations using Unicode NFC normalisation and diacritic-insensitive scoring. NFC standardises equivalent Unicode representations, while the latter removes combining marks to assess their effect on reference-based evaluation. Both transformations are applied to predictions and references.

\begin{table}[ht!]
\centering
\scriptsize
\setlength{\tabcolsep}{4pt}
\begin{tabular}{|l|l|r|r|r|}
\hline
\textbf{Model} &
\textbf{Evaluation} &
\textbf{BLEU $\uparrow$} &
\textbf{chrF $\uparrow$} &
\textbf{TER $\downarrow$} \\
\hline

\multirow{3}{*}{GPT-5.1}
& Original & 4.06 & 19.70 & 98.41 \\
& NFC & 6.11 & 29.11 & 93.39 \\
& Diacritic-insensitive & \textbf{13.47} & \textbf{42.64} & \textbf{81.46} \\
\hline

\multirow{3}{*}{Llama-70B}
& Original & 1.95 & 17.50 & 94.32 \\
& NFC & 2.99 & 25.55 & 90.79 \\
& Diacritic-insensitive & \textbf{7.50} & \textbf{36.82} & \textbf{82.77} \\
\hline

\multirow{3}{*}{Gemma-27B}
& Original & 1.07 & 14.49 & 97.14 \\
& NFC & 2.88 & 25.27 & 90.76 \\
& Diacritic-insensitive & \textbf{7.35} & \textbf{36.80} & \textbf{82.53} \\
\hline
\end{tabular}
\caption{Effect of orthographic normalisation on ReciLex-augmented English-to-Yor\`ub\'a translation, averaged across zero-shot and few-shot prompting.}
\label{tab:yoruba_normalisation_ablation}
\end{table}

Table~\ref{tab:yoruba_normalisation_ablation} shows substantial improvements in ReciLex scores following orthographic normalisation. NFC consistently improves translation scores, while diacritic-insensitive evaluation produces larger gains across all three models. These findings indicate that orthographic variation contributes substantially to the lower reference-based scores initially observed for ReciLex.

\subsection{Comparison with Dedicated Machine Translation Models}
\label{sec:ablation_mt_comparison}
We compare zero-shot translation using GPT-5.1, Llama-70B, and Gemma-27B, with and without ReciLex, against two dedicated MT models: M2M100-1.2B and NLLB-200-600M. Results are averaged across Yor\`ub\'a, Hausa, and Igbo, with LLM results additionally averaged across the three models.

Tables~\ref{tab:llm_mt_auto_comparison} and \ref{tab:llm_mt_human_comparison} show that zero-shot LLMs generally outperform both dedicated MT models across automatic and human evaluation metrics. ReciLex augmentation further improves LLM performance, achieving the highest scores across all metrics and cultural adaptation dimensions. Although NLLB-200 consistently outperforms M2M100, both models fall behind the ReciLex-augmented LLMs.

\begin{table}[ht!]
\centering
\scriptsize
\setlength{\tabcolsep}{4pt}
\begin{tabular}{|l|r|r|r|}
\hline
\textbf{System} &
\textbf{BLEU $\uparrow$} &
\textbf{chrF $\uparrow$} &
\textbf{TER $\downarrow$} \\
\hline
LLMs (Zero-shot)
& 6.36 & 29.80 & 84.93 \\
LLMs (Zero-shot + ReciLex)
& \textbf{10.70} & \textbf{34.96} & \textbf{78.78} \\
M2M100-1.2B
& 0.90 & 13.57 & 96.20 \\
NLLB-200-600M
& 6.83 & 26.43 & 89.47 \\
\hline
\end{tabular}
\caption{Mean automatic evaluation scores across Yor\`ub\'a, Hausa, and Igbo. LLM results are averaged across GPT-5.1, Llama-70B, and Gemma-27B using zero-shot prompting. Bold indicates the better score.}
\label{tab:llm_mt_auto_comparison}
\end{table}

\begin{table}[ht!]
\centering
\scriptsize
\setlength{\tabcolsep}{3pt}
\begin{tabular}{|l|r|r|r|r|r|}
\hline
\textbf{System} &
\textbf{INGR} &
\textbf{PROC} &
\textbf{SEM} &
\textbf{CULT} &
\textbf{LING} \\
\hline
LLMs (Zero-shot)
& 2.76 & 2.84 & 2.84 & 2.42 & 2.49 \\
LLMs (Zero-shot + ReciLex)
& \textbf{3.30} & \textbf{3.11}
& \textbf{3.07} & \textbf{2.78}
& \textbf{2.85} \\
M2M100-1.2B
& 1.66 & 1.58 & 1.43 & 1.33 & 1.19 \\
NLLB-200-600M
& 2.66 & 2.57 & 2.50 & 2.22 & 2.19 \\
\hline
\end{tabular}
\caption{Mean human cultural adaptation scores across three target languages. LLM results are averaged across GPT-5.1, Llama-70B, and Gemma-27B using zero-shot prompting. Bold indicates the better score.}
\label{tab:llm_mt_human_comparison}
\end{table}

\subsection{Impact of Zero-Shot and Few-Shot Prompting}
\label{sec:ablation_prompting}
To investigate the impact of in-context examples, we compare zero-shot and few-shot prompting across GPT-5.1, Llama-70B, and Gemma-27B for English-to-Yor\`ub\'a, Hausa, and Igbo recipe translation. Results are aggregated across models, target languages, and ReciLex configurations to isolate the overall effect of prompting strategy.

Tables~\ref{tab:prompting_auto_ablation} and
\ref{tab:prompting_human_ablation} show that few-shot prompting substantially outperforms zero-shot prompting across all automatic evaluation metrics, increasing BLEU and chrF, while reducing TER. Human evaluation similarly shows improvements across all five cultural adaptation dimensions, particularly ingredient localisation and linguistic naturalness. These findings demonstrate the overall effectiveness of in-context examples in improving both translation quality and cultural adaptation, regardless of whether lexical augmentation is applied.

\begin{table}[ht!]
\centering
\scriptsize
\setlength{\tabcolsep}{5pt}
\begin{tabular}{|l|r|r|r|}
\hline
\textbf{Setting} &
\textbf{BLEU $\uparrow$} &
\textbf{chrF $\uparrow$} &
\textbf{TER $\downarrow$} \\
\hline
Zero-shot & 8.53 & 32.38 & 81.86 \\
Few-shot & \textbf{19.55} & \textbf{44.26} &
\textbf{70.41} \\
\hline
\end{tabular}
\caption{Impact of prompting strategy on automatic recipe translation performance, averaged across three LLMs, three target languages, and both ReciLex configurations. Bold indicates the better score.}
\label{tab:prompting_auto_ablation}
\end{table}

\begin{table}[ht!]
\centering
\scriptsize
\setlength{\tabcolsep}{3pt}
\begin{tabular}{|l|r|r|r|r|r|}
\hline
\textbf{Setting} &
\textbf{INGR} &
\textbf{PROC} &
\textbf{SEM} &
\textbf{CULT} &
\textbf{LING} \\
\hline
Zero-shot
& 3.03 & 2.98 & 2.96 & 2.60 & 2.67 \\
Few-shot
& \textbf{3.36} & \textbf{3.19} &
\textbf{3.13} & \textbf{2.82} &
\textbf{2.96} \\
\hline
\end{tabular}
\caption{Impact of prompting strategy on human cultural adaptation ratings, averaged across three LLMs, three target languages, and both ReciLex configurations. Bold indicates the better score.}
\label{tab:prompting_human_ablation}
\end{table}

\subsection{Prompt Language Sensitivity}
\label{sec:ablation_prompt_language}
To test whether prompt language affects translation performance, we compare English-language prompts with prompts written in the corresponding target language (Yor\`ub\'a, Hausa, or Igbo). Results are aggregated across GPT-5.1, Llama-70B, and Gemma-27B, including zero-shot and few-shot prompting with and without ReciLex. This isolates prompt language while averaging over the other experimental configurations.

Tables~\ref{tab:prompt_language_auto} and \ref{tab:prompt_language_human} show that English-language prompting yields modest improvements in automatic translation performance, increasing BLEU and chrF, while reducing TER. Human evaluation similarly indicates improvements in ingredient localisation, procedural localisation, and linguistic naturalness. However, semantic preservation remains nearly unchanged, while cultural authenticity decreases slightly. Based on these observations, we adopt English-language prompting for the main experiments reported in this work.

\begin{table}[ht!]
\centering
\small
\setlength{\tabcolsep}{5pt}
\begin{tabular}{|l|r|r|r|}
\hline
\textbf{Prompt Language} &
\textbf{BLEU $\uparrow$} &
\textbf{chrF $\uparrow$} &
\textbf{TER $\downarrow$} \\
\hline
Target LRL & 14.04 & 38.32 & 76.13 \\
English & \textbf{15.02} & \textbf{39.61}
& \textbf{75.74} \\
\hline
\end{tabular}
\caption{Automatic translation performance under English- and target-language prompting, averaged across three LLMs, three target languages, zero-shot and few-shot prompting, and both ReciLex configurations. Bold indicates the better score.}
\label{tab:prompt_language_auto}
\end{table}

\begin{table}[ht!]
\centering
\small
\setlength{\tabcolsep}{3pt}
\begin{tabular}{|l|r|r|r|r|r|}
\hline
\textbf{Prompt} &
\textbf{INGR} &
\textbf{PROC} &
\textbf{SEM} &
\textbf{CULT} &
\textbf{LING} \\
\hline

Target LRL & 3.18 & 3.11 & \textbf{3.06}
& \textbf{2.77} & 2.83 \\
English
& \textbf{3.22} & \textbf{3.15}
& 3.06 & 2.76 & \textbf{2.92} \\
\hline
\end{tabular}
\caption{Human cultural adaptation ratings for English- and target-language prompting, based on matched translation outputs. Results are averaged across three LLMs, three target languages, and prompting configurations. Bold indicates the better score.}
\label{tab:prompt_language_human}
\end{table}

\subsection{Effect of Few-Shot Example Selection}
\label{appendix:fewshot_comparison}
Table~\ref{tab:fewshot_comparison} compares the results of 3-shot, 5-shot, and 10-shot prompting across the evaluated translation settings and languages. Results show that 5-shot prompting consistently improves performance over 3-shot prompting across most metrics and translation directions, while remaining more stable and computationally efficient than larger prompting configurations. Based on these observations, we adopt 5-shot prompting for the main experiments reported in this work.

\begin{table}[ht!]
\centering
\scriptsize
\setlength{\tabcolsep}{5pt}
\begin{tabular}{|l|l|c|c|c|}
\hline
\textbf{Language} &
\textbf{Shots} &
\textbf{BLEU $\uparrow$} &
\textbf{chrF $\uparrow$} &
\textbf{TER $\downarrow$} \\
\hline

\multirow{3}{*}{Yor\`ub\'a}
& 3-shot  & 13.39 & 38.26 & 82.65 \\
& 5-shot  & \textbf{16.51} & \textbf{41.13} & \textbf{78.64} \\
& 10-shot & \underline{16.28} & \underline{39.31} & \underline{79.52} \\
\hline

\multirow{3}{*}{Hausa}
& 3-shot  & \underline{20.09} & \underline{50.91} & 72.39 \\
& 5-shot  & 19.83 & 49.65 & \textbf{69.87} \\
& 10-shot & \textbf{20.33} & \textbf{51.66} & \underline{71.34} \\
\hline

\multirow{3}{*}{Igbo}
& 3-shot  & 24.84 & 50.83 & 67.82 \\
& 5-shot  & \textbf{27.09} & \textbf{53.41} & \textbf{63.92} \\
& 10-shot & \underline{26.44} & \underline{52.23} & \underline{65.25} \\
\hline

\end{tabular}
\caption{Comparison of 3-shot, 5-shot, and 10-shot prompting using GPT-5.1 for translation across Yor\`ub\'a, Hausa, and Igbo. Experiments use English prompts and 50 ReciLoReL samples. Bold and underlined indicate the best and second-best scores, respectively.}
\label{tab:fewshot_comparison}
\end{table}

\section{Human Cultural Adaptation Evaluation}
\subsection{Human Annotation Coverage}
\label{sec:annotation_coverage}
Table~\ref{tab:annotation_distribution} summarises the distribution of manually evaluated recipe translations across the experimental conditions, including LLM translations using English and target-language prompts, and translations produced by dedicated MT models. Counts represent unique evaluated outputs, excluding duplicate IAA judgements.

\begin{table}[t]
\centering
\scriptsize
\setlength{\tabcolsep}{2pt}
\begin{tabular}{|l|l|r|}
\hline
\textbf{Category} & \textbf{Configuration} & \textbf{Total} \\
\hline
\multicolumn{3}{|l|}{\textbf{Overall annotation coverage}} \\
\hline
Source recipes & Unique recipes & 80 \\
Translated recipes & All configurations & 3,900 \\
\hline

\multicolumn{3}{|l|}{\textbf{Translation system}} \\
\hline
\multirow{2}{*}{System}
& LLM translations & 3,420 \\
& MT translations & 480 \\
\hline

\multicolumn{3}{|l|}{\textbf{Target language}} \\
\hline
\multirow{3}{*}{Language}
& Yor\`ub\'a & 1,300 \\
& Hausa & 1,300 \\
& Igbo & 1,300 \\
\hline

\multicolumn{3}{|l|}{\textbf{Language model}} \\
\hline
\multirow{3}{*}{LLM}
& GPT-5.1 & 1,140 \\
& Llama-70B & 1,140 \\
& Gemma-27B & 1,140 \\
\hline

\multicolumn{3}{|l|}{\textbf{Dedicated MT models}} \\
\hline
\multirow{2}{*}{MT}
& M2M100-1.2B & 240 \\
& NLLB-200-600M & 240 \\
\hline

\multicolumn{3}{|l|}{\textbf{Prompting strategy (LLMs)}} \\
\hline
\multirow{2}{*}{Strategy}
& Zero-shot & 1,710 \\
& Few-shot & 1,710 \\
\hline

\multicolumn{3}{|l|}{\textbf{Lexicon augmentation (LLMs)}} \\
\hline
\multirow{2}{*}{ReciLex}
& Without ReciLex & 1,710 \\
& With ReciLex & 1,710 \\
\hline

\multicolumn{3}{|l|}{\textbf{Prompt language (LLMs)}} \\
\hline
\multirow{2}{*}{Prompt}
& English & 2,880 \\
& Target language & 540 \\
\hline

\multicolumn{3}{|l|}{\textbf{Per-language breakdown}} \\
\hline
LLM & English prompts & 960 \\
LLM & Target-language prompts & 180 \\
MT & Both MT models & 160 \\
\hline

\multicolumn{3}{|l|}{\textbf{Individual experimental configurations}} \\
\hline
LLM & English prompt configuration & 80 \\
LLM & Target-language prompt configuration & 15 \\
MT & Each model per language & 80 \\
\hline

\end{tabular}
\caption{Distribution of manually evaluated recipe translations across target languages, translation systems, prompting strategies, lexicon augmentation conditions, and prompt languages. Counts represent unique translated recipe outputs, excluding duplicate IAA judgements.}
\label{tab:annotation_distribution}
\end{table}

\subsection{Inter-Annotator Agreement}
\label{app:cultural_agreement}
Two annotators evaluated each language, independently rating 80 shared translated recipes, while the remaining samples were assigned individually. Inter-annotator agreement (IAA) was measured using graded soft agreement, which accounts for rating differences on the five-point evaluation scale by assigning partial credit rather than requiring exact agreement. Scores range from 0 to 1, with higher values indicating closer agreement. Identical ratings receive full credit (1.0), while differences of one, two, three, and four points receive 0.75, 0.50, 0.25, and 0.0, respectively.

\begin{table}[ht!]
\centering
\scriptsize
\setlength{\tabcolsep}{5pt}
\begin{tabular}{|l|c|c|c|}
\hline
\textbf{Evaluation Dimension} &
\textbf{Yor\`ub\'a} &
\textbf{Hausa} &
\textbf{Igbo} \\
\hline
Ingredient localisation & 0.631 & 0.747 & 0.831 \\
Procedural localisation & 0.644 & 0.697 & 0.816 \\
Semantic preservation & 0.556 & 0.694 & 0.844 \\
Cultural authenticity & 0.650 & 0.766 & 0.878 \\
Linguistic naturalness & 0.669 & 0.772 & 0.881 \\
\hline
\textbf{Average} &
\textbf{0.630} &
\textbf{0.735} &
\textbf{0.850} \\
\hline
\end{tabular}
\caption{Inter-annotator agreement measured using graded soft agreement across five cultural adaptation dimensions. Higher scores indicate closer agreement.}
\label{tab:cultural_agreement}
\end{table}

Table~\ref{tab:cultural_agreement} reports agreement across the five evaluation dimensions. Igbo achieved the highest average agreement (0.850), followed by Hausa (0.735) and Yor\`ub\'a (0.630). Under the distance-weighted measure, these correspond to average absolute rating differences of 0.60, 1.06, and 1.48 points, respectively.

\section{Qualitative Error Analysis}
\label{sec:qualitative-error-analysis}
We examine GPT-5.1 translations of the same 80 recipes across Hausa, Igbo, and Yor\`ub\'a under zero-shot and few-shot prompting, with and without ReciLex. From 474 matched comparisons, we select ten illustrative cases based on human-evaluation scores and observable translation differences. Table~\ref{tab:qualitative-errors} presents five cases covering all five cultural adaptation evaluation criteria.

As shown in Table~\ref{tab:qualitative-errors}, ReciLex can improve ingredient localisation and linguistic naturalness by providing more appropriate culinary terminology, particularly where baseline translations retain English expressions or use incorrect lexical equivalents. However, these improvements do not guarantee semantic or procedural fidelity. The incorrect ingredient and cooking instructions observed in Hausa suggest that lexical guidance may introduce or amplify errors when terms are selected without sufficient contextual disambiguation or when their use affects the surrounding instruction.

\begin{table*}[ht!]
\centering
\scriptsize
\setlength{\tabcolsep}{2.5pt}
\renewcommand{\arraystretch}{1.12}

\begin{tabular}{|p{2.0cm}|p{2.7cm}|p{6.0cm}|c|p{3.6cm}|}
\hline

\textbf{Recipe Details} &
\textbf{English Source} &
\textbf{Translation} &
\textbf{CA Criterion} &
\textbf{Error Analysis} \\
\hline

Chicken Casserole (Hausa, Few-Shot) &

Mix soup and milk together; stir until smooth. Fold in sour cream and stir until smooth. &

\textbf{ReciLex:} \emph{Ka caku\texthtd{}a Miya da Madara tare; ka motsa har sai ta zama lallausa. Sai ka lankafa Madara mai tsami ka motsa har sai ta zama lallausa.}

\textbf{Without ReciLex:} \emph{Ka ha\texthtd{}a miya (soup) da madara a wuri guda; ka juya har su yi laushi su zama \texthtd{}aya. Ka \texthtk{}ara sour cream ka juya har ya yi laushi sosai.} &

INGR &

ReciLex replaces the untranslated \emph{sour cream} with the Hausa description \emph{Madara mai tsami}. This improves lexical localisation, although the expression is less specific than the original ingredient. \\
\hline

Fish A La Creme (Igbo, Zero-Shot) &

Marinate fish in lemon juice for 10 minutes. Mix butter, soup and sour cream. Pour over fish. &

\textbf{ReciLex:} \emph{Tee ngwa nri az\d{u} n'ime mman\d{u} lemuni (lemon juice) maka nkeji 10. Gwak\d{o} b\d{o}ta, supu na sour cream. Gbanye ya n'elu az\d{u}.}

\textbf{Without ReciLex:} \emph{Tinye an\d{u} \d{o}h\d{i}a n'ime mmanya \d{o}s\d{i}sa lemon ka \d{o} mara na ya nkeji 10. Kp\d{o}\d{o} \d{o}n\d{u} ma k\d{u}\d{o} butter, soup na sour cream.} &

SEM &

The baseline incorrectly translates \emph{fish} as \emph{an\d{u} \d{o}h\d{i}a} (wild animal). ReciLex preserves the intended ingredient using \emph{az\d{u}} (fish), although its rendering of lemon juice remains semantically inaccurate. \\
\hline

Hotel Roanoke Spoon Bread (Igbo, Zero-Shot) &

Combine water, butter, sugar and salt in a saucepan; bring to a boil. Stir in cornmeal and cook until thickened. &

\textbf{ReciLex:} \emph{Gwak\d{o}ta mmiri, b\d{o}ta, shuga na nnu n'ite s\d{o}\d{o}s\d{u} nta; ruo \d{o}s\d{u}s\d{u}. Gbanye cornmeal ma gb\d{o}\d{o} agb\d{o} Nkeji otu.}

\textbf{Without ReciLex:} \emph{Tinye mmiri, butter, sugar na salt n'ime obere saucepan; k\d{u}daa ya ruo mgbe \d{o} ga-ab\d{u} boil. Tinye cornmeal ma k\d{u}daa ya.} &

LING &

The baseline retains several English terms, including \emph{butter}, \emph{sugar}, \emph{salt}, and \emph{boil}. ReciLex provides more localised lexical forms, reducing code-mixing, although \emph{cornmeal} remains untranslated. \\
\hline

Shrimp Fried Rice (Yor\`ub\'a, Few-Shot) &

Add green onions, garlic, shrimp and peas. Stir-fry for 2 minutes. &

\textbf{ReciLex:} \emph{Fi \`{a}l\`{u}b\'{\d{o}}s\`{a} el\'{e}w\'{e}, \`{a}l\`{u}b\'{o}s\`{a} \'{a}\'{a}y\`{u}, ed\'{e} \`{a}ti \`{\d{e}}p\`{a} kun un. Yan l\'{a}s\'{a}n f\'{u}n \`{i}\d{s}\'{\d{e}}j\'{u} m\'{e}j\`{i}.}

\textbf{Without ReciLex:} \emph{Fi alubosa \`{a}t\`{a}r\d{e} al\'{a}w\`{\d{o}} ew\'{e}, allar\'{o} (garlic), shrimpu \`{a}ti \`{e}p\`{a} (peas) s\'{i} i. Stir-fry f\'{u}n \`{i}\d{s}\'{\d{e}}j\'{u} m\'{e}j\`{i}.} &

CULT &

ReciLex replaces the English-derived \emph{shrimpu} with \emph{ed\'{e}}, a more recognisable Yor\`ub\'a culinary term. However, its rendering of \emph{stir-fry} is less procedurally precise. \\
\hline

Fish A La Creme (Hausa, Few-Shot) &

Marinate fish in lemon juice for 10 minutes. Mix butter, soup and sour cream. Pour over fish. Broil for 10 minutes. &

\textbf{ReciLex:} \emph{Ka tsoma a ruwa Kifi a cikin Kunun lemun tsami na tsawon da\texthtk{}i\texthtk{}a goma. Ka caku\texthtd{}a Man shanu, Miya da Madara mai tsami. Ka zuba a kan Kifi. Ka tafasa na da\texthtk{}i\texthtk{}a goma.}

\textbf{Without ReciLex:} \emph{Ka jika kifin da ruwan lemun tsami na tsawon minti goma. Ka ha\texthtd{}a bota, miya (soup) da madarar sour cream. Ka zuba a kan kifin. Ka soya a \texthtk{}ar\texthtk{}ashin broiler na tsawon minti goma.} &

PROC &

ReciLex changes both ten-minute durations to ten seconds and replaces \emph{broiling} with \emph{boiling}. Despite localising some ingredient terms, it alters the cooking method and duration, making the instructions unreliable. \\
\hline
\end{tabular}
\caption{Illustrative GPT-5.1 recipe translation improvements and errors with and without ReciLex.}
\label{tab:qualitative-errors}
\end{table*}

\section{ReciLoRel Recipe Selection Process}
\label{appendix:recilorel_recipe_selection}
Algorithm~\ref{algorithm:recipe_selection_algorithm} summarises the procedure for identifying and selecting recipes for expert translation and cultural adaptation in ReciLoReL. We first used the ReciFine entity annotations to identify recipes containing entities from all ten culinary entity categories (see Appendix~\ref{appendix:recifine_et_cats}). This ensured that selected recipes contain diverse ingredients, cooking tools, quantities, durations, food states, and procedural cooking actions necessary for culturally adaptive translation evaluation. 

We also filtered recipes with extremely short or excessively long ingredient lists and cooking instructions to obtain recipes with moderate procedural complexity and lexical diversity. From the resulting filtered recipe pool, we randomly sampled 200 recipes for expert translation and cultural adaptation into Yor\`ub\'a, Hausa, and Igbo.

\begin{algorithm}[ht]
\caption{Recipe Selection for ReciLoReL}
\scriptsize
\begin{algorithmic}[1]
\label{algorithm:recipe_selection_algorithm}
\STATE Extract recipe entities using ReciFine annotations
\STATE Load entities for each of the 10 entity categories
\FOR{each recipe in RecipeNLG}
    \STATE Check whether recipe contains at least one entity from all entity type categories
    \STATE Filter recipes with extreme instruction or ingredient lengths
\ENDFOR
\STATE Randomly sample 200 recipes from the filtered pool
\end{algorithmic}
\end{algorithm}

\section{ReciLoRel Descriptive Statistics}
\label{appendix:desc_stats_recilorel}
Table \ref{tab:recilorel_dataset_statistics} presents descriptive statistics for ReciLoReL across English, Yor\`ub\'a, Hausa, and Igbo. The table compares lexical and structural properties of the translated recipes, including average title and recipe lengths, vocabulary diversity, and translation expansion ratios relative to the English source recipes. We additionally report annotation statistics, including the number of annotators and expert reviewers involved in the construction of the benchmark resources.

\begin{table}[ht!]
\centering
\scriptsize
\resizebox{\columnwidth}{!}{
\begin{tabular}{|l|c|c|c|c|}
\hline
\textbf{Statistic} & \textbf{English} & \textbf{Yor\`ub\'a} & \textbf{Hausa} & \textbf{Igbo} \\
\hline

Recipes & 200 & 200 & 200 & 200 \\
\hline

Avg. title words & 3.10 & 3.80 & 3.80 & 3.90 \\
Avg. title characters & 20.90 & 23.10 & 21.90 & 22.90 \\
Unique title words & 295 & 391 & 254 & 280 \\
\hline

Avg. recipe words & 68.20 & 108.60 & 103.20 & 104.00 \\
Avg. recipe characters & 383.90 & 503.90 & 524.50 & 525.60 \\
Avg. recipe sentences & 8.50 & 8.40 & 9.00 & 9.30 \\
Unique recipe words & 995 & 1,980 & 947 & 1,050 \\
Lexical diversity & 0.77 & 0.69 & 0.63 & 0.64 \\
Word expansion ratio & 1.00 & 1.61 & 1.53 & 1.55 \\
Character expansion ratio & 1.00 & 1.33 & 1.38 & 1.39 \\
\hline

Annotators & -- & 5 & 2 & 2 \\
Expert reviewers & -- & 3 & 3 & 3 \\
Avg. annotation time (hr.) & -- & 1.50 & 0.90 & 1.00 \\
\hline

\end{tabular}
}
\caption{Descriptive statistics for ReciLoReL across English, Yor\`ub\'a, Hausa, and Igbo. Expansion ratios are computed relative to the English source recipes.}
\label{tab:recilorel_dataset_statistics}
\end{table}

\section{ReciLex Descriptive Statistics}
\label{appendix:desc_stats_recilex}
Table~\ref{tab:recilex_statistics} presents descriptive statistics for ReciLex across Yor\`ub\'a, Hausa, and Igbo. The table summarises the overall scale and lexical characteristics of the multilingual culinary lexicons, including the total number of translated entries, covered entity categories, average entry lengths, vocabulary sizes, and lexical diversity measures. We additionally report word and character expansion ratios relative to the English source lexicons, highlighting the tendency of translated culinary expressions in the target languages to be longer than their English counterparts. We also report the number of annotators and the number of expert reviewers together with the average time spent annotating each lexicon entry.

\begin{table}[ht!]
\centering
\scriptsize
\resizebox{\columnwidth}{!}{
\begin{tabular}{|l|c|c|c|c|}
\hline
\textbf{Statistic} & \textbf{English} & \textbf{Yor\`ub\'a} & \textbf{Hausa} & \textbf{Igbo} \\
\hline

Lexicon entries & 5000 & 5000 & 5000 & 5000 \\
Entity categories & 10 & 10 & 10 & 10 \\
\hline

Avg. words & 1.60 & 2.90 & 2.30 & 2.40 \\
Avg. characters & 7.90 & 17.60 & 11.80 & 12.20 \\
Unique words & 1,789 & 2,294 & 1,324 & 1,839 \\
Lexical diversity & 0.22 & 0.16 & 0.11 & 0.18 \\
\hline

Word expansion ratio & 1.00 & 1.97 & 1.55 & 1.54 \\
Character expansion ratio & 1.00 & 2.60 & 1.64 & 1.68 \\
\hline

Annotators & -- & 5 & 2 & 2 \\
Expert reviewers & -- & 3 & 3 & 3 \\
Avg. time per entry (min.) & -- & 2.50 & 0.80 & 1.50 \\
\hline

\end{tabular}
}
\caption{Descriptive statistics for ReciLex across English, Yor\`ub\'a, Hausa, and Igbo. Expansion ratios are computed relative to the English source lexicon.}
\label{tab:recilex_statistics}
\end{table}

\section{ReciFine Entity Categories}
\label{appendix:recifine_et_cats}
Table~\ref{tab:erfg-entity-tags} presents the ten culinary entity categories adopted from ReciFine, which provides automatically extracted culinary entities from over 2.2 million recipes. We use these extracted entities to identify the most frequent entities within each category and construct ReciLex for multilingual culinary lexicon translation and culturally adaptive recipe generation.

\begin{table}[ht!]
\centering
\scriptsize
\setlength{\tabcolsep}{3pt}
\renewcommand{\arraystretch}{1.15}
\begin{tabular}{p{0.08\columnwidth} p{0.25\columnwidth} p{0.60\columnwidth}}
\hline
\textbf{Tag} & \textbf{Name} & \textbf{Definition} \\
\hline
F   & Food             & Edible items, including both raw ingredients and intermediate products \\
\hline
T   & Tool             & Cooking tools such as \textit{knives, bowls, pans} \\
\hline
D   & Duration         & Time durations in cooking (\textit{20 minutes}) \\
\hline
Q   & Quantity         & Quantities associated with ingredients \\
\hline
Ac  & Action by chef   & Verbs for deliberate cook actions (\textit{bring} in “Bring the mixture to a boil”) \\
\hline
Ac2 & Discontinuous Ac & Non-contiguous parts of compound chef actions (\textit{to a boil}) \\
\hline
Af  & Action by food   & Verbs where food is the agent (\textit{melt, boil}) \\
\hline
At  & Action by tool   & Verbs denoting tool actions (\textit{grind, beat}) \\
\hline
Sf  & Food state       & Descriptions of food’s state (\textit{chopped, soft}) \\
\hline
St  & Tool state       & Descriptions of tool readiness (\textit{preheated, greased, covered}) \\
\hline
\end{tabular}
\caption{Entity types extracted in the ReciFine corpus.}
\label{tab:erfg-entity-tags}
\end{table}

\begin{table*}[ht!]
\centering
\scriptsize
\setlength{\tabcolsep}{5pt}
\renewcommand{\arraystretch}{1.15}
\resizebox{\textwidth}{!}{
\begin{tabular}{|l|c|c|l|c|c|c|}
\hline
\textbf{Model} &
\textbf{Params} &
\textbf{Type} &
\textbf{Inference Backend} &
\textbf{Quant.} &
\textbf{Max. Input Tokens} &
\textbf{Access} \\
\hline
GPT-5.1 & Unknown & Proprietary & OpenAI API & -- & 2048 & API \\
\hline
Llama-3.3-70B & 70B & Open-weight & Unsloth & 4-bit & 2048 & Local/HF \\
\hline
Gemma-27B & 27B & Open-weight & Unsloth & 4-bit & 2048 & Local/HF \\
\hline
NLLB-200 & 600M & Multilingual MT & HuggingFace Transformers & -- & 2048 & Local/HF \\
\hline
M2M-100-1.2B & 1.2B & Multilingual MT & HuggingFace Transformers & -- & 2048 & Local/HF \\
\hline
\end{tabular}
}
\caption{
Overview of the six instruction-tuned LLMs and two multilingual MT baselines evaluated in this work. LLMs are evaluated using prompting-only inference without parameter fine-tuning, while MT models perform direct translation. All models support evaluation on Hausa, Yorùbá, and Igbo.
}
\label{tab:model_configurations}
\end{table*}

\section{LLMs and MT Models Evaluated}
\label{app:model_configurations}
We evaluate six multilingual instruction-tuned large language models: GPT-5.1, Llama-3.3-70B, and Gemma-27B. These models were selected due to their strong multilingual capabilities and recent performance on multilingual reasoning and generation tasks. All open-weight models were evaluated using the Unsloth inference framework for efficient long-context generation and memory-efficient inference. GPT-5.1 was accessed through the OpenAI API. Since our experiments focus on zero-shot and few-shot culturally adaptive recipe translation, no parameter finetuning was performed for any model.

We additionally evaluate NLLB-200 \citep{fan2021beyond} and M2M-100~\citep{fan2021beyond} as dedicated multilingual machine translation (MT) baselines to provide a reference for direct translation performance. All three LLMs were evaluated across all prompting conditions and target languages (Hausa, Yorùbá, and Igbo). We evaluated the two MT models, NLLB-200 and M2M-100, on all three languages. Further details on the model architectures, parameter sizes, inference backends, quantisation settings, and access methods are provided in Table~\ref{tab:model_configurations}.

\section{Dataset Annotation Guideline}
\label{appendix:annotation_guideline}
This section presents the annotation guidelines used for constructing ReciLoReL and ReciLex, as well as the guidelines used for the human evaluation of cultural adaptation. For resource construction, annotators were instructed to produce translations that are semantically faithful to the source text while remaining culturally appropriate for target-language communities. Translations were expected to adapt ingredients, cooking terminology, measurements, utensils, and cooking procedures as needed, while preserving the recipe's original intent. We additionally provide the guidelines used to evaluate generated recipe translations based on the five cultural adaptation criteria defined in Section~\ref{sec:cultural-evaluation}.

\subsection{ReciLoReL Recipe Translation Guideline}
\label{appendix:recilorel_recipe_translation_guideline}
\subsubsection{Task Overview}
Annotators were asked to translate complete recipe titles and cooking instructions from English into the target low-resource language (Hausa, Yor\`ub\'a, or Igbo). Unlike isolated texts, recipe translation provides full contextual information, including ingredients, preparation steps, cooking actions, and sequencing. The primary goal of ReciLoReL was to create culturally adaptive multilingual recipe translations that remain faithful to the source recipe while sounding natural and usable for target-language speakers.

\subsubsection{General Translation Principles}
All recipe translations were required to follow the principles below:

\begin{itemize}
    \item Preserve the original meaning and cooking intent of the recipe.
    \item Maintain ingredient names, quantities, temperatures, and cooking actions whenever possible.
    \item Produce fluent and natural translations suitable for real cooking instructions.
    \item Avoid overly literal translations that sound unnatural in the target language.
    \item Do not summarise, omit, or simplify recipe steps.
    \item Ensure that translated recipes remain cookable and practically understandable.
\end{itemize}

\subsubsection{Cultural Adaptation Requirements}
Annotators were additionally encouraged to adapt translations to the target culinary culture where appropriate. This includes:

\begin{itemize}
    \item Using culturally familiar ingredient names or locally recognised equivalents.
    \item Adapting borrowed culinary terms into natural target-language forms where commonly used.
    \item Using culturally familiar cooking expressions and preparation terminology.
    \item Preserving local readability and usability rather than performing rigid word-for-word translation.
\end{itemize}

For example, ingredients or tools without widely used native equivalents could either retain their English form or be adapted into a locally naturalised pronunciation or spelling convention.

\subsubsection{Handling Ambiguity}
When recipes contained ambiguous culinary expressions, annotators were instructed to prioritise translations that remained natural and understandable in practical cooking contexts. Cultural appropriateness and usability were prioritised over literal lexical correspondence.

\subsection{ReciLex Lexicon Translation Guideline}
\label{appendix:recilex_lexicon_translation_guideline}
\subsubsection{Task Overview}
Annotators were also asked to translate isolated culinary lexical items and short phrases extracted from recipes. These lexicons were grouped into entity categories derived from the ReciFine corpus (see Appendix~\ref{appendix:recifine_et_cats}). Unlike full recipes, these entries were presented without context sentences, requiring annotators to infer their most appropriate culinary interpretation. The translated lexicons were later reintegrated into recipe prompting pipelines for culturally adaptive multilingual recipe generation.

\subsubsection{General Translation Principles}
All lexicon translations were required to follow the principles below:

\begin{itemize}
    \item Translations must be concise, natural, and suitable for cooking instructions.
    \item Prefer commonly used culinary terminology over literal or uncommon phrasing.
    \item Maintain the original semantic role of the lexical item.
    \item Avoid unnecessary explanations or descriptive additions.
    \item Ensure that translated entries can fit naturally into complete recipe instructions.
\end{itemize}

\subsubsection{Cultural Adaptation Requirements}
Annotators were instructed to adapt lexical translations to the target culinary culture whenever necessary. This includes:

\begin{itemize}
    \item Using culturally familiar ingredient names and cooking terminology.
    \item Adapting foreign food names into locally recognisable forms when appropriate.
    \item Preferring translations that would sound natural to native speakers in practical cooking settings.
    \item Ensuring compatibility with real recipe usage in the target language community.
\end{itemize}

\subsubsection{ReciLex Entity Categories}
The ReciLex annotations were organised into 10 categories as in ReciFine (see Appendix~\ref{appendix:recifine_et_cats}). Each category captures a distinct functional component of cooking instructions and recipe semantics.

\subsubsection{Handling Ambiguity}
Because lexical items were presented in isolation:

\begin{itemize}
    \item Some entries may appear ambiguous without sentence context.
    \item Annotators were instructed to use culinary knowledge and common cooking conventions to infer the most appropriate translation.
    \item Preference was given to translations that would remain suitable across multiple recipe contexts rather than a single narrow interpretation.
\end{itemize}

\subsection{Cultural Adaptation Evaluation Guideline}
\label{appendix:cultural_evaluation_guideline}
Human annotators were asked to evaluate generated recipe translations using the five cultural adaptation criteria defined in Section~\ref{sec:cultural-evaluation} without being shown the gold reference. For each generated recipe, annotators were provided with the source recipe and the generated translation and asked to independently assign a score from 1 (very poor) to 5 (very good) for each criterion.

Annotators were instructed that cultural adaptation does not require every element of a recipe to be modified. Ingredients, procedures, measurements, and equipment that are already appropriate for the target context should be preserved, while adaptation should be considered where direct transfer would make an element unfamiliar, inaccessible, unnatural, or impractical. Any adaptation should remain consistent with the meaning and culinary intent of the source recipe.

The following questions were provided to guide the evaluation:

\begin{itemize}
    \item Ingredient Localisation: Are the ingredients familiar, culturally appropriate, and reasonably available in the target context?

    \item Procedural Localisation: Are the cooking methods, equipment, and measurements appropriate and practical in the target context?

    \item Semantic Preservation: Does the adapted recipe preserve the intended culinary meaning and outcome of the source recipe?

    \item Cultural Authenticity: Does the adapted recipe represent a culturally plausible recipe within the target context?

    \item Linguistic Naturalness: Does the recipe read naturally using culinary language and conventions of the target language?
\end{itemize}

Scores were assigned using the following scale:

\begin{itemize}
    \item 5 -- Very Good: Fully satisfies the evaluation criterion,
    with no meaningful issues.
    \item 4 -- Good: Satisfies the criterion with only minor issues.
    \item 3 -- Acceptable: Generally satisfies the criterion, but
    contains noticeable issues.
    \item 2 -- Poor: Only partially satisfies the criterion, with
    substantial issues.
    \item 1 -- Very Poor: Does not satisfy the criterion or contains
    major problems.
\end{itemize}

Annotators were instructed to evaluate the five criteria independently, as a recipe may perform well on one aspect of cultural adaptation while performing poorly on another. Table~\ref{tab:cultural-dimensions}

\begin{table*}[ht!]
\centering
\scriptsize
\resizebox{\textwidth}{!}{
\begin{tabular}{|l|l|}
\hline
\textbf{Evaluation Criteria} & \textbf{Evaluation Focus} \\
\hline

Ingredient Localisation &
Are the ingredients familiar, culturally appropriate, and reasonably available in the target context? \\
\hline

Procedural Localisation &
Are the cooking methods, equipment, and measurements appropriate for the target context? \\
\hline

Semantic Preservation &
Does the adapted recipe preserve the intended culinary meaning and outcome of the source recipe? \\
\hline

Cultural Authenticity &
Does the adapted recipe represent a culturally plausible recipe within the target context? \\
\hline

Linguistic Naturalness &
Does the recipe read naturally using culinary language and conventions of the target language? \\
\hline

\end{tabular}
}
\caption{Criteria for evaluating cultural adaptation in recipe translation.}
\label{tab:cultural-dimensions}
\end{table*}

\section{Prompt Template for Recipe Cultural Adaptation}
This section presents the prompting templates used for culturally adaptive recipe translation. We evaluate four prompting configurations: zero-shot prompting without ReciLex augmentation, few-shot prompting without ReciLex augmentation, zero-shot prompting with ReciLex augmentation, and few-shot prompting with ReciLex augmentation. For all settings, prompts were written either in English or directly in the target low-resource language. For clarity, we present representative prompting examples from a single language for each prompting configuration. The zero-shot prompting templates without ReciLex augmentation in Section~\ref{appendix:zero-shot-no-recilex} use Yor\`ub\'a examples. The few-shot prompting templates without ReciLex augmentation in Section~\ref{appendix:few-shot-no-recilex} use Hausa examples. The zero-shot prompting templates with ReciLex augmentation in Section~\ref{appendix:zero-shot-with-recilex} use Igbo examples. Finally, the few-shot prompting templates with ReciLex augmentation in Section~\ref{appendix:few-shot-with-recilex} use Hausa examples.

\subsection{Zero-Shot Prompting without ReciLex}
\label{appendix:zero-shot-no-recilex}
For zero-shot prompting without ReciLex augmentation, we construct prompts in four languages: English, Hausa, Yor\`ub\'a, and Igbo. This lets us evaluate whether prompting directly in the target low-resource language improves culturally adaptive recipe translation compared to prompts written in English. Figures~\ref{fig:zero_shot_en_to_yo_en_prompt}, \ref{fig:zero_shot_en_to_yo_yo_prompt}, \ref{fig:zero_shot_yo_to_en_en_prompt}, and \ref{fig:zero_shot_yo_to_en_yo_prompt} present the four Yor\`ub\'a-related zero-shot prompting templates used in our experiments, covering both English-to-Yor\`ub\'a and Yor\`ub\'a-to-English translation using either English or Yor\`ub\'a prompt instructions. Prompt templates for Hausa and Igbo follow the same overall structure, differing only in the target language used in the prompt instructions and translation outputs.

\begin{figure}[ht!]
\footnotesize
\fbox{
\parbox{0.94\columnwidth}{
\sloppy
\textbf{You are an expert culinary translator with strong knowledge of recipes, ingredients, cooking actions, food terminology, and culturally adaptive cooking practices.}

Translate the following recipe title and recipe text in their entirety from English to Yoruba (a language spoken in Western Nigeria).

Ensure that the translated recipe is culturally natural and relevant for Yoruba speakers by considering locally familiar ingredients, cooking tools, preparation methods, and culturally appropriate culinary expressions where suitable.

If no direct Yoruba equivalent exists, keep the original English term or adapt it into a Yoruba phonological form where appropriate. For instance, ``rice'' may be rendered as \`{\i}r\`{\d{e}}s\`{\i}, ``bread'' as b\'ur\`{\d{e}}d\`{\i}, and ``sugar'' as \d{s}\'ug\`a.

\textbf{Requirements:}
\begin{itemize}
\item Preserve the original meaning exactly.
\item Preserve ingredient names, quantities, temperatures, and cooking actions.
\item Maintain cultural appropriateness and naturalness for Yoruba speakers.
\item Do not summarise.
\item Do not explain.
\item Return your answer strictly in valid JSON.
\end{itemize}

\texttt{\{"translated\_title": "<Yoruba translation of the title>",}

\texttt{"translated\_recipe": "<Yoruba translation of the recipe>"\}}

\textbf{Title:} <English recipe title>

\textbf{Recipe:} <English recipe text>
}
}
\caption{English-to-Yor\`ub\'a zero-shot prompting using English instructions.}
\label{fig:zero_shot_en_to_yo_en_prompt}
\end{figure}

\begin{figure}[ht!]
\footnotesize
\fbox{
\parbox{0.94\columnwidth}{
\sloppy
\textbf{\`Iwo j\'e am\`oye n\'inu \`itum\`o\ onj\d{e} \`ati \`it\'um\`o\ \`ed\`e, p\d{e}l\'u \`im\`o\ jinl\d{e} n\'ipa \`awon \`il\`ana onj\d{e}, \`awon \`er\`oja, \`awon \`i\d{s}e \`id\'an\'a, \`awon \`or\`o \`im\`o\ onj\d{e}, \`ati \`im\'ul\`o \`ib\'il\d{e} n\'inu \`id\'an\'a.}

T\'u ak\`ol\'e ohunelo \`ati gbogbo \`ak\'o\'on\'u ohunelo t\'o w\`a l\'or\'i s\'i \`ed\`e Yor\`ub\'a l\'ati \`ed\`e G\`{\d{e}}\'{\d{e}}s\`{\i}.

J\d{e} k\'i ohunelo n\'a\`a d\'ara f\'un \`awon ol\`us\d{o} Yor\`ub\'a n\'ipa l\'ilo \`awon \`er\`oja, irin\d{s}\d{e} \`id\'an\'a, \`ati \`on\`a \`id\'an\'a t\'i w\d{o}n m\d{o} s\'i, \`ati \`awon \`or\`o onj\d{e} t\'o b\'a \`a\d{s}\`a Yor\`ub\'a mu.

T\'i k\`o b\'a s\'i \`or\`o Yor\`ub\'a t\'o d\'aj\'u f\'un \`or\`o kan, fi \`or\`o G\`{\d{e}}\'{\d{e}}s\`{\i} n\'a\`a s\'il\d{e} g\d{e}g\d{e} b\'i \'o \d{s}e w\`a, t\'ab\'i \d{s}e \`at\'un\d{s}e r\d{e} s\'i ir\'u \`ip\`e Yor\`ub\'a t\'o b\'a y\d{e}.

\textbf{\`Awon \`il\`ana:}
\begin{itemize}
\item Pa \`itum\`o\ gangan m\d{o}.
\item Pa \`awon or\'uk\d{o} \`er\`oja, iye, \`igb\'on\'a, \`ati \`awon \`i\d{s}e \`id\'an\'a m\d{o}.
\item J\d{e} k\'i \`itum\`o\ n\'a\`a b\'a \`a\d{s}\`a \`ati \`im\'ul\`o Yor\`ub\'a mu.
\item M\'a \d{s}e k\'ur\`o n\'i \`ak\'o\'on\'u t\'ab\'i k\'ur\`o n\'inu \`it\'um\`o.
\item M\'a \d{s}e \d{s}\`al\`ay\'e.
\item D\'ah\`un r\d{e} gb\d{o}d\d{o} w\`a n\'inu JSON t\'o p\'eye.
\end{itemize}

\texttt{\{"translated\_title": "<\`Itum\`o\ Yor\`ub\'a ti ak\`ol\'e>",}

\texttt{"translated\_recipe": "<\`Itum\`o\ Yor\`ub\'a ti ohunelo>"\}}

\textbf{Ak\`ol\'e:} <English recipe title>

\textbf{Ohunelo:} <English recipe text>
}
}
\caption{English-to-Yor\`ub\'a zero-shot prompting using Yor\`ub\'a instructions.}
\label{fig:zero_shot_en_to_yo_yo_prompt}
\end{figure}

\begin{figure}[t]
\footnotesize
\fbox{
\parbox{0.94\columnwidth}{
\sloppy
\textbf{You are an expert culinary translator with strong knowledge of recipes, ingredients, cooking actions, food terminology, and culturally adaptive cooking practices.}

Translate the following recipe title and recipe text in their entirety from Yoruba (a language spoken in Western Nigeria) into English.

Ensure that the translated recipe remains culturally faithful while also being fluent, natural, and understandable to English speakers.

If a direct English equivalent for a word is unclear, provide the closest natural English rendering while preserving the original meaning.

\textbf{Requirements:}
\begin{itemize}
\item Preserve the original meaning exactly.
\item Preserve ingredient names, quantities, temperatures, and cooking actions.
\item Preserve culturally important culinary concepts where appropriate.
\item Do not summarise.
\item Do not explain.
\item Return your answer strictly in valid JSON.
\end{itemize}

\texttt{\{"translated\_title": "<English translation of the title>",}

\texttt{"translated\_recipe": "<English translation of the recipe>"\}}

\textbf{Title:} <Yoruba recipe title>

\textbf{Recipe:} <Yoruba recipe text>
}
}
\caption{Yor\`ub\'a-to-English zero-shot prompting using English instructions.}
\label{fig:zero_shot_yo_to_en_en_prompt}
\end{figure}

\begin{figure}[ht!]
\footnotesize
\fbox{
\parbox{0.94\columnwidth}{
\sloppy
\textbf{\`Iwo j\'e am\`oye n\'inu \`itum\`o\ onj\d{e} \`ati \`it\'um\`o\ \`ed\`e, p\d{e}l\'u \`im\`o\ jinl\d{e} n\'ipa \`awon ohunelo, \`awon \`er\`oja, \`awon \`i\d{s}e \`id\'an\'a, \`ati \`awon \`or\`o \`im\`o\ onj\d{e}.}

T\'u ak\`ol\'e ohunelo \`ati gbogbo \`ak\'o\'on\'u ohunelo l\'ati \`ed\`e Yor\`ub\'a s\'i \`ed\`e G\`{\d{e}}\'{\d{e}}s\`{\i}.

J\d{e} k\'i \`itum\`o\ n\'a\`a ba \`a\d{s}\`a \`ati \`im\`o\ onj\d{e} Yor\`ub\'a m\d{o}, \d{s}\`ugb\'on k\'i \'o t\'un j\'e m\'imu, adayeba, \`ati kedere n\'inu \`ed\`e G\`{\d{e}}\'{\d{e}}s\`{\i}.

T\'i \`itum\`o\ G\`{\d{e}}\'{\d{e}}s\`{\i} t\'o p\'eye f\'un \`or\`o kan k\`o b\'a h\`an s\`{\i} gbangba, fi \`itum\`o\ G\`{\d{e}}\'{\d{e}}s\`{\i} t\'o s\'unm\d{o} t\'i \'o s\`i d\'ara l\'or\'i \`ed\`e h\`an, \d{s}\`ugb\'on pa \`itum\`o\ \`ak\d{o}k\d{o} m\d{o}.

\textbf{\`Awon \`il\`ana:}
\begin{itemize}
\item Pa \`itum\`o\ gangan m\d{o}.
\item Pa \`awon or\'uk\d{o} \`er\`oja, iye, \`igb\'on\'a, \`ati \`awon \`i\d{s}e \`id\'an\'a m\d{o}.
\item Pa \`awon \`afih\`an \`a\d{s}\`a onj\d{e} Yor\`ub\'a t\'o \d{s}e p\`at\`ak\`i m\d{o}.
\item M\'a \d{s}e k\'ur\`o n\'i \`ak\'o\'on\'u t\'ab\'i d\'in oro k\`u.
\item M\'a \d{s}e \d{s}\`al\`ay\'e.
\item D\'ah\`un r\d{e} gb\d{o}d\d{o} w\`a n\'inu JSON t\'o p\'eye.
\end{itemize}

\texttt{\{"translated\_title": "<\`Itum\`o\ G\`{\d{e}}\'{\d{e}}s\`{\i} ti ak\`ol\'e>",}

\texttt{"translated\_recipe": "<\`Itum\`o\ G\`{\d{e}}\'{\d{e}}s\`{\i} ti ohunelo>"\}}

\textbf{Ak\`ol\'e:} <Yoruba recipe title>

\textbf{Ohunelo:} <Yoruba recipe text>
}
}
\caption{Yor\`ub\'a-to-English zero-shot prompting using Yor\`ub\'a instructions.}
\label{fig:zero_shot_yo_to_en_yo_prompt}
\end{figure}

\subsection{Few-Shot Prompting without ReciLex}
\label{appendix:few-shot-no-recilex}

For few-shot prompting without ReciLex augmentation, we prepend five expert-translated recipe examples from the support split before the test recipe. This enables the model to observe the expected output format, translation style, and cultural adaptation behaviour before generating the target translation. Figures~\ref{fig:few_shot_en_to_ha_en_prompt}, \ref{fig:few_shot_en_to_ha_ha_prompt}, \ref{fig:few_shot_ha_to_en_en_prompt}, and \ref{fig:few_shot_ha_to_en_ha_prompt} show the Hausa-related few-shot prompting templates. Equivalent templates were constructed for Yor\`ub\'a and Igbo by replacing the target language and few-shot examples.

\begin{figure}[ht!]
\footnotesize
\fbox{
\parbox{0.94\columnwidth}{
\sloppy
\textbf{You are an expert culinary translator with strong knowledge of recipes, ingredients, cooking actions, food terminology, and culturally adaptive cooking practices.}

Translate the following recipe title and recipe text in their entirety from English to Hausa.

Ensure that the translated recipe is culturally natural and relevant for Hausa speakers by considering locally familiar ingredients, cooking tools, preparation methods, and culturally appropriate culinary expressions where suitable.

If a direct Hausa equivalent for a word is unavailable, retain the original English term or adapt it into a Hausa phonological form where appropriate.

\textbf{Requirements:}
\begin{itemize}
\item Preserve the original meaning exactly.
\item Preserve ingredient names, quantities, temperatures, and cooking actions.
\item Maintain cultural appropriateness and naturalness for Hausa speakers.
\item Do not summarise.
\item Do not explain.
\item Return your answer strictly in valid JSON.
\end{itemize}

\texttt{\{"translated\_title": "<Hausa translation of the title>",}

\texttt{"translated\_recipe": "<Hausa translation of the recipe>"\}}

Below are example English-to-Hausa recipe translations.

\textbf{Example 1--5:}\\
Title: <English example title>\\
Recipe: <English example recipe>\\
Response: \texttt{\{"translated\_title": "...", "translated\_recipe": "..."\}}

Now translate the following recipe.

\textbf{Title:} <English recipe title>

\textbf{Recipe:} <English recipe text>
}
}
\caption{English-to-Hausa few-shot prompting using English instructions.}
\label{fig:few_shot_en_to_ha_en_prompt}
\end{figure}

\begin{figure}[ht!]
\footnotesize
\fbox{
\parbox{0.94\columnwidth}{
\sloppy
\textbf{Kai \texthtk wararren mai fassarar girke-girke ne, kana da \texthtk warewa kan girke-girke, kayan abinci, ayyukan girki, kalmomin abinci, da daidaita girki da al'adar masu amfani da harshe.}

Fassara taken girke-girke da cikakken rubutun girke-girke daga Turanci zuwa Hausa.

Ka tabbatar cewa fassarar ta dace da al'adar Hausawa ta hanyar la'akari da kayan abinci, kayan girki, hanyoyin sarrafa abinci, da kalmomin girki da suka saba a cikin al'ummar Hausa.

Idan babu madaidaicin kalmar Hausa ga wata kalma, ka bar kalmar Turanci yadda take ko ka daidaita ta da lafazin Hausa idan ya dace.

\textbf{Ka'idoji:}
\begin{itemize}
\item Ka kiyaye ainihin ma'ana.
\item Ka kiyaye sunayen kayan abinci, ma'auni, da ayyukan girki.
\item Ka tabbatar fassarar ta dace da al'adar Hausawa.
\item Kada ka ta\texthtk aita.
\item Kada ka yi bayani.
\item Ka mayar da amsa cikin JSON mai inganci kawai.
\end{itemize}

\texttt{\{"translated\_title": "<Fassarar taken zuwa Hausa>",}

\texttt{"translated\_recipe": "<Fassarar girke-girke zuwa Hausa>"\}}

A \texthtk asa akwai misalan fassarar girke-girke daga Turanci zuwa Hausa.

\textbf{Misali 1--5:}\\
Take: <English example title>\\
Girke-girke: <English example recipe>\\
Amsa: \texttt{\{"translated\_title": "...", "translated\_recipe": "..."\}}

Yanzu fassara girke-girken nan.

\textbf{Take:} <English recipe title>

\textbf{Girke-girke:} <English recipe text>
}
}
\caption{English-to-Hausa few-shot prompting using Hausa instructions.}
\label{fig:few_shot_en_to_ha_ha_prompt}
\end{figure}

\begin{figure}[ht!]
\footnotesize
\fbox{
\parbox{0.94\columnwidth}{
\sloppy
\textbf{You are an expert culinary translator with strong knowledge of recipes, ingredients, cooking actions, food terminology, and culturally adaptive cooking practices.}

Translate the following recipe title and recipe text in their entirety from Hausa into English.

Ensure that the translated recipe remains culturally faithful while also being fluent, natural, and understandable to English speakers.

If a direct English equivalent for a word is unclear, provide the closest natural English rendering while preserving the original meaning.

\textbf{Requirements:}
\begin{itemize}
\item Preserve the original meaning exactly.
\item Preserve ingredient names, quantities, temperatures, and cooking actions.
\item Preserve culturally important culinary concepts where appropriate.
\item Do not summarise.
\item Do not explain.
\item Return your answer strictly in valid JSON.
\end{itemize}

\texttt{\{"translated\_title": "<English translation of the title>",}

\texttt{"translated\_recipe": "<English translation of the recipe>"\}}

Below are example Hausa-to-English recipe translations.

\textbf{Example 1--5:}\\
Title: <Hausa example title>\\
Recipe: <Hausa example recipe>\\
Response: \texttt{\{"translated\_title": "...", "translated\_recipe": "..."\}}

Now translate the following recipe.

\textbf{Title:} <Hausa recipe title>

\textbf{Recipe:} <Hausa recipe text>
}
}
\caption{Hausa-to-English few-shot prompting using English instructions.}
\label{fig:few_shot_ha_to_en_en_prompt}
\end{figure}

\begin{figure}[ht!]
\footnotesize
\fbox{
\parbox{0.94\columnwidth}{
\sloppy
\textbf{Kai \texthtk wararren mai fassarar girke-girke ne, kana da \texthtk warewa kan girke-girke, kayan abinci, ayyukan girki, da kalmomin abinci.}

Fassara taken girke-girke da cikakken rubutun girke-girke daga Hausa zuwa Turanci.

Ka tabbatar cewa fassarar ta ri\texthtk e ma'anar al'adar abincin Hausa, amma kuma ta kasance mai kyau, ta halitta, kuma a bayyane cikin Turanci.

Idan madaidaicin fassarar Turanci ga wata kalma ba ta bayyana ba, ka yi amfani da fassarar Turanci mafi kusa da ma'anarta.

\textbf{Ka'idoji:}
\begin{itemize}
\item Ka kiyaye ainihin ma'ana.
\item Ka kiyaye sunayen kayan abinci, ma'auni, da ayyukan girki.
\item Ka kiyaye muhimman ra'ayoyin al'adar abinci.
\item Kada ka ta\texthtk aita.
\item Kada ka yi bayani.
\item Ka mayar da amsa cikin JSON mai inganci kawai.
\end{itemize}

\texttt{\{"translated\_title": "<Fassarar taken zuwa Turanci>",}

\texttt{"translated\_recipe": "<Fassarar girke-girke zuwa Turanci>"\}}

A \texthtk asa akwai misalan fassarar girke-girke daga Hausa zuwa Turanci.

\textbf{Misali 1--5:}\\
Take: <Hausa example title>\\
Girke-girke: <Hausa example recipe>\\
Amsa: \texttt{\{"translated\_title": "...", "translated\_recipe": "..."\}}

Yanzu fassara girke-girken nan zuwa Turanci.

\textbf{Take:} <Hausa recipe title>

\textbf{Girke-girke:} <Hausa recipe text>
}
}
\caption{Hausa-to-English few-shot prompting using Hausa instructions.}
\label{fig:few_shot_ha_to_en_ha_prompt}
\end{figure}

\subsection{Zero-Shot Prompting ReciLex}
\label{appendix:zero-shot-with-recilex}
For zero-shot prompting ReciLex augmentation, we retrieve recipe-specific culinary entities from the ReciFine annotations and provide their translated ReciLex entries as additional lexical guidance. We group the retrieved lexicon by entity category and insert it into the prompt before the recipe to be translated. The model is instructed to use the provided translations where appropriate, while avoiding forced substitutions that would make the output unnatural. Figures~\ref{fig:zero_shot_recilex_en_to_ig_en_prompt}, \ref{fig:zero_shot_recilex_en_to_ig_ig_prompt}, \ref{fig:zero_shot_recilex_ig_to_en_en_prompt}, and \ref{fig:zero_shot_recilex_ig_to_en_ig_prompt} show representative Igbo ReciLex zero-shot prompting templates. Templates for Yor\`ub\'a and Hausa follow the same structure.

\begin{figure}[ht!]
\footnotesize
\fbox{
\parbox{0.94\columnwidth}{
\sloppy
\textbf{You are an expert culinary translator with strong knowledge of recipes, ingredients, cooking actions, and food terminology.}

Translate the following recipe title and recipe text in their entirety from English to Igbo.

You are also given a partial recipe lexicon containing some relevant English culinary entities and their Igbo translations. This lexicon is not exhaustive and may not cover every word in the recipe. Use the provided translations where they are appropriate, but maintain fluent and natural Igbo.

\textbf{Partial relevant lexicon:}\\
<entity-type grouped ReciLex entries>

If a direct Igbo equivalent for a word is unavailable, retain the original English term or adapt it into a natural Igbo-compatible form where appropriate.

\textbf{Requirements:}
\begin{itemize}
\item Preserve the original meaning exactly.
\item Preserve ingredient names, quantities, temperatures, and cooking actions.
\item Use the provided lexicon translations where appropriate.
\item Do not force a lexicon translation if it makes the Igbo unnatural or incorrect.
\item Do not summarise.
\item Do not explain.
\item Return your answer strictly in valid JSON.
\end{itemize}

\texttt{\{"translated\_title": "", "translated\_recipe": ""\}}

\textbf{Title:} <English recipe title>

\textbf{Recipe:} <English recipe text>
}
}
\caption{English-to-Igbo zero-shot ReciLex prompting using English instructions.}
\label{fig:zero_shot_recilex_en_to_ig_en_prompt}
\end{figure}

\begin{figure}[ht!]
\footnotesize
\fbox{
\parbox{0.94\columnwidth}{
\sloppy
\textbf{\d{I} b\d{u} \d{o}kachamara n'\d{i}t\d{u}ghar\d{i} as\d{u}s\d{u} nri nwere ezigbo \d{o}m\d{u}ma gbasara nri, ihe eji esi nri, usoro esi nri, na okwu met\d{u}tara uzommeputa nri.}

T\d{u}ghar\d{i}a aha uzommeputa nri na nk\d{o}wa uzommeputa niile d\d{i} n'okpuru site na Bekee gaa na Igbo.

A nyekwara g\d{i} obere lexicon nwere \d{u}f\d{o}d\d{u} okwu nri Bekee na ihe eji esi nri, yana nt\d{u}ghar\d{i} ha gaa na Igbo. Lexicon a ezugh\d{i} oke, o nwekwara ike \d{o} gagh\d{i} ekpuchi okwu niile d\d{i} n'uzommeputa ah\d{u}. Jiri nt\d{u}ghar\d{i} enyere ebe \d{o} dabara, mana gbaa mb\d{o} h\d{u} na Igbo ah\d{u} d\d{i} mfe ngh\d{o}ta, d\d{i} n'usoro, ma b\d{u}r\d{u} nke ziri ezi.

\textbf{Lexicon dabara na uzommeputa a:}\\
<entity-type grouped ReciLex entries>

\d{O} b\d{u}r\d{u} na enwegh\d{i} ezigbo okwu Igbo kwek\d{o}r\d{o} na otu okwu Bekee, hap\d{u} okwu Bekee ah\d{u} d\d{i}ka \d{o} d\d{i} ma \d{o} b\d{u} gbanwee ya, ka \d{o} dabara nke \d{o}ma n'Igbo.

\textbf{Ntuziaka:}
\begin{itemize}
\item Debe ihe okwu mb\d{u} p\d{u}tara nke \d{o}ma.
\item Debe aha ihe eji esi nri, \d{o}n\d{u}\d{o}g\d{u}, okpom\d{o}k\d{u}, na omume eji esi nri.
\item Jiri nt\d{u}ghar\d{i} lexicon enyere ebe \d{o} dabara.
\item Ekwela ka \d{i} manye nt\d{u}ghar\d{i} lexicon ma \d{o} b\d{u}r\d{u} na \d{o} ga-eme ka Igbo ah\d{u} ghara \d{i}d\d{i} mma.
\item Ekwela ka nk\d{o}wa ah\d{u} d\d{i} mkp\d{u}mkp\d{u} ma \d{o} b\d{u} wep\d{u} ozi d\d{i} mkpa.
\item Ekwela nk\d{o}wa mgbakwunye.
\item Weghachite naan\d{i} az\d{i}za g\d{i} n'usoro JSON ziri ezi.
\end{itemize}

\texttt{\{"translated\_title": "", "translated\_recipe": ""\}}

\textbf{Aha:} <English recipe title>

\textbf{Uzommeputa:} <English recipe text>
}
}
\caption{English-to-Igbo zero-shot ReciLex prompting using Igbo instructions.}
\label{fig:zero_shot_recilex_en_to_ig_ig_prompt}
\end{figure}

\begin{figure}[ht!]
\footnotesize
\fbox{
\parbox{0.94\columnwidth}{
\sloppy
\textbf{You are an expert culinary translator with strong knowledge of recipes, ingredients, cooking actions, and food terminology.}

Translate the following recipe title and recipe text in their entirety from Igbo into English.

You are also given a partial recipe lexicon containing some relevant Igbo culinary entities and their English translations. This lexicon is not exhaustive and may not cover every word in the recipe. Use the provided translations where they are appropriate, but maintain fluent and natural English.

\textbf{Partial relevant lexicon:}\\
<entity-type grouped ReciLex entries>

If a direct English equivalent for a word is unclear, provide the closest natural English rendering while preserving the original meaning.

\textbf{Requirements:}
\begin{itemize}
\item Preserve the original meaning exactly.
\item Preserve ingredient names, quantities, temperatures, and cooking actions.
\item Use the provided lexicon translations where appropriate.
\item Do not force a lexicon translation if it makes the English unnatural or incorrect.
\item Do not summarise.
\item Do not explain.
\item Return your answer strictly in valid JSON.
\end{itemize}

\texttt{\{"translated\_title": "", "translated\_recipe": ""\}}

\textbf{Title:} <Igbo recipe title>

\textbf{Recipe:} <Igbo recipe text>
}
}
\caption{Igbo-to-English zero-shot ReciLex prompting using English instructions.}
\label{fig:zero_shot_recilex_ig_to_en_en_prompt}
\end{figure}

\begin{figure}[ht!]
\footnotesize
\fbox{
\parbox{0.94\columnwidth}{
\sloppy
\textbf{\d{I} b\d{u} \d{o}kachamara n'\d{i}t\d{u}ghar\d{i} as\d{u}s\d{u} nri nwere ezigbo \d{o}m\d{u}ma gbasara uzommeputa nri, ihe eji esi nri, omume eji esi nri, na okwu met\d{u}tara nri.}

T\d{u}ghar\d{i}a aha uzommeputa nri na nk\d{o}wa uzommeputa niile d\d{i} n'okpuru site na Igbo gaa na Bekee.

A nyekwara g\d{i} obere lexicon nwere \d{u}f\d{o}d\d{u} okwu nri Igbo na ihe eji esi nri, yana nt\d{u}ghar\d{i} ha gaa na Bekee. Lexicon a ezugh\d{i} oke, o nwekwara ike \d{o} gagh\d{i} ekpuchi okwu niile d\d{i} n'uzommeputa ah\d{u}. Jiri nt\d{u}ghar\d{i} enyere ebe \d{o} dabara, mana gbaa mb\d{o} h\d{u} na Bekee ah\d{u} d\d{i} mfe ngh\d{o}ta, d\d{i} n'usoro, ma b\d{u}r\d{u} nke ziri ezi.

\textbf{Lexicon dabara na uzommeputa a:}\\
<entity-type grouped ReciLex entries>

\d{O} b\d{u}r\d{u} na ezigbo okwu Bekee kwek\d{o}r\d{o} na otu okwu Igbo ap\d{u}tagh\d{i} nke \d{o}ma, nye nt\d{u}ghar\d{i} Bekee kacha nso nke na-echekwa ihe okwu mb\d{u} p\d{u}tara.

\textbf{Ntuziaka:}
\begin{itemize}
\item Debe ihe okwu mb\d{u} p\d{u}tara nke \d{o}ma.
\item Debe aha ihe eji esi nri, \d{o}n\d{u}\d{o}g\d{u}, okpom\d{o}k\d{u}, na omume eji esi nri.
\item Jiri nt\d{u}ghar\d{i} lexicon enyere ebe \d{o} dabara.
\item Ekwela ka \d{i} manye nt\d{u}ghar\d{i} lexicon ma \d{o} b\d{u}r\d{u} na \d{o} ga-eme ka Bekee ah\d{u} ghara \d{i}d\d{i} mma.
\item Ekwela ka nk\d{o}wa ah\d{u} d\d{i} mkp\d{u}mkp\d{u} ma \d{o} b\d{u} belata ya.
\item Ekwela nk\d{o}wa mgbakwunye.
\item Weghachite naan\d{i} az\d{i}za g\d{i} n'usoro JSON ziri ezi.
\end{itemize}

\texttt{\{"translated\_title": "", "translated\_recipe": ""\}}

\textbf{Aha:} <Igbo recipe title>

\textbf{Uzommeputa:} <Igbo recipe text>
}
}
\caption{Igbo-to-English zero-shot ReciLex prompting using Igbo instructions.}
\label{fig:zero_shot_recilex_ig_to_en_ig_prompt}
\end{figure}

\subsection{Few-Shot Prompting ReciLex}
\label{appendix:few-shot-with-recilex}

For few-shot prompting ReciLex augmentation, we combine recipe-specific ReciLex lexical guidance with multiple in-context translation demonstrations. Similar to the zero-shot ReciLex setting, recipe-specific culinary entities are retrieved from the ReciFine annotations and inserted into the prompt as grouped lexical translation pairs. In addition, the prompt includes five example recipe translations that demonstrate the expected translation style and JSON output structure. Figures~\ref{fig:fewshot_recilex_en_ha_en}, \ref{fig:fewshot_recilex_en_ha_ha}, \ref{fig:fewshot_recilex_ha_en_en}, and \ref{fig:fewshot_recilex_ha_en_ha} present representative Hausa few-shot ReciLex prompting templates. Prompt variants for Yor\`ub\'a and Igbo follow the same structure.

\begin{figure}[t]
\footnotesize
\fbox{
\parbox{0.95\columnwidth}{
\sloppy
\textbf{You are an expert culinary translator with strong knowledge of recipes, ingredients, cooking actions, and food terminology.}

Translate the following recipe title and recipe text in their entirety from English to Hausa.

\textbf{Partial relevant lexicon:}\\
<entity-type grouped ReciLex entries>

\textbf{Requirements:}
\begin{itemize}
\item Preserve the original meaning exactly.
\item Preserve ingredient names, quantities, temperatures, and cooking actions.
\item Use the provided lexicon translations where appropriate.
\item Do not summarise.
\item Return valid JSON only.
\end{itemize}

\textbf{Example 1--5:}\\
Title: <English example title>\\
Recipe: <English example recipe>\\
Response: \texttt{\{"translated\_title": "...", "translated\_recipe": "..."\}}

Now translate the following recipe into Hausa.

\textbf{Title:} <English recipe title>

\textbf{Recipe:} <English recipe text>
}
}
\caption{English-to-Hausa few-shot ReciLex prompting using English instructions.}
\label{fig:fewshot_recilex_en_ha_en}
\end{figure}

\begin{figure}[t]
\footnotesize
\fbox{
\parbox{0.95\columnwidth}{
\sloppy
\textbf{Kai \texthtk wararren mai fassarar girke-girke ne tare da cikakken ilimi kan abinci, kayan ha i, hanyoyin girki, da kalmomin da suka shafi girke-girke.}

Fassara taken girke-girken da cikakken bayanin girke-girken da ke \texthtk asa daga Turanci zuwa Hausa.

\textbf{Lexicon da ya dace da wannan girke-girke:}\\
<entity-type grouped ReciLex entries>

\textbf{Ka’idoji:}
\begin{itemize}
\item Ka kiyaye cikakkiyar ma’ana ta asali.
\item Ka kiyaye sunayen kayan ha\texthtd i, ma’auni, yanayin zafi, da hanyoyin girki.
\item Yi amfani da fassarorin lexicon da aka bayar inda suka dace.
\item Kada ka ta\texthtk aita bayani.
\item Ka dawo da amsar cikin ingantaccen JSON kawai.
\end{itemize}

\textbf{Misali 1--5:}\\
Take: <English example title>\\
Girke-girke: <English example recipe>\\
Amsa: \texttt{\{"translated\_title": "...", "translated\_recipe": "..."\}}

Yanzu fassara girke-girken nan zuwa Hausa.

\textbf{Take:} <English recipe title>

\textbf{Girke-girke:} <English recipe text>
}
}
\caption{English-to-Hausa few-shot ReciLex prompting using Hausa instructions.}
\label{fig:fewshot_recilex_en_ha_ha}
\end{figure}

\begin{figure}[t]
\footnotesize
\fbox{
\parbox{0.95\columnwidth}{
\sloppy
\textbf{You are an expert culinary translator with strong knowledge of recipes, ingredients, cooking actions, and food terminology.}

Translate the following recipe title and recipe text in their entirety from Hausa into English.

\textbf{Partial relevant lexicon:}\\
<entity-type grouped ReciLex entries>

\textbf{Requirements:}
\begin{itemize}
\item Preserve the original meaning exactly.
\item Preserve ingredient names, quantities, temperatures, and cooking actions.
\item Use the provided lexicon translations where appropriate.
\item Do not summarise.
\item Return valid JSON only.
\end{itemize}

\textbf{Example 1--5:}\\
Title: <Hausa example title>\\
Recipe: <Hausa example recipe>\\
Response: \texttt{\{"translated\_title": "...", "translated\_recipe": "..."\}}

Now translate the following recipe into English.

\textbf{Title:} <Hausa recipe title>

\textbf{Recipe:} <Hausa recipe text>
}
}
\caption{Hausa-to-English few-shot ReciLex prompting using English instructions.}
\label{fig:fewshot_recilex_ha_en_en}
\end{figure}

\begin{figure}[t]
\footnotesize
\fbox{
\parbox{0.95\columnwidth}{
\sloppy
\textbf{Kai \texthtk wararren mai fassarar girke-girke ne tare da cikakken ilimi kan girke-girke, kayan ha\texthtd i, hanyoyin girki, da kalmomin da suka shafi abinci.}

Fassara taken girke-girken da cikakken bayanin girke-girken da ke \texthtk asa daga Hausa zuwa Turanci.

\textbf{Lexicon da ya dace da wannan girke-girke:}\\
<entity-type grouped ReciLex entries>

\textbf{Ka’idoji:}
\begin{itemize}
\item Ka kiyaye cikakkiyar ma’ana ta asali.
\item Ka kiyaye sunayen kayan ha\texthtd i, ma’auni, yanayin zafi, da hanyoyin girki.
\item Yi amfani da fassarorin lexicon da aka bayar inda suka dace.
\item Kada ka rage bayani.
\item Ka dawo da amsar cikin ingantaccen JSON kawai.
\end{itemize}

\textbf{Misali 1--5:}\\
Take: <Hausa example title>\\
Girke-girke: <Hausa example recipe>\\
Amsa: \texttt{\{"translated\_title": "...", "translated\_recipe": "..."\}}

Yanzu fassara girke-girken nan zuwa Turanci.

\textbf{Take:} <Hausa recipe title>

\textbf{Girke-girke:} <Hausa recipe text>
}
}
\caption{Hausa-to-English few-shot ReciLex prompting using Hausa instructions.}
\label{fig:fewshot_recilex_ha_en_ha}
\end{figure}
\end{document}